\documentclass[journal]{IEEEtran}
\usepackage{times}
\usepackage{epsfig}
\usepackage{graphicx}
\usepackage{subfigure}
\usepackage{amsmath}
\usepackage{mathtools}
\usepackage{amssymb}
\usepackage[noblocks]{authblk}
\usepackage[bookmarks=false]{hyperref}
\usepackage{bm}
\usepackage{color}
\usepackage{multirow}
\usepackage{geometry}

\newtheorem{thm}{\noindent \textbf{Theorem}}

\DeclarePairedDelimiter{\norm}{\lVert}{\rVert}

\usepackage{cite}

\ifCLASSINFOpdf
\else
\fi

\def\degree{${}^{\circ}$}
\newcommand{\modifyA}[1]{{\color{black}#1}}
\newcommand{\modifyB}[1]{{\color{black}#1}}
\newcommand{\modifyC}[1]{{\color{black}#1}}
\newcommand{\modifyD}[1]{{\color{black}#1}}
\newcommand{\modifyE}[1]{{\color{black}#1}}
\newcommand{\modifyF}[1]{{\color{black}#1}}
\newcommand{\modifyG}[1]{{\color{black}#1}}
\newcommand{\tabincell}[2]{
\begin{tabular}{@{}#1@{}}#2\end{tabular}
}
\newcommand{\erhao}{\fontsize{21pt}{\baselineskip}\selectfont}

\begin{document}
%

\newgeometry{top=6cm,bottom=1cm}

\onecolumn{

\noindent \textbf{\erhao{Robust Depth-based Person Re-identification}}

\vspace{2cm}

\noindent {\LARGE{Ancong Wu, Wei-Shi Zheng, Jian-Huang Lai}}

\Large
\vspace{2cm}

\noindent Code is available at the project page: \\
\ \ \ \ \ \ \ \ \ \ \ \ \url{http://isee.sysu.edu.cn/~wuancong/ProjectDepthReID.htm}

\vspace{1cm}

\noindent For reference of this work, please cite:

\vspace{1cm}
\noindent Ancong Wu, Wei-Shi Zheng, Jian-Huang Lai. Robust Depth-based Person Re-identification. IEEE Transactions on Image Processing (DOI:10.1109/TIP.2017.2675201)

\vspace{1cm}

\noindent Bib:\\
\noindent
@article\{wu2017depth,\\
\ \ \   title=\{Robust Depth-based Person Re-identification\},\\
\ \ \  author=\{Wu, Ancong and Zheng, Wei-Shi and Lai, Jianhuang\},\\
\ \ \  journal=\{IEEE Transactions on Image Processing\\(DOI:10.1109/TIP.2017.2675201)\},\\
\ \ \  year=\{2017\}\\
\}

}

\clearpage

\restoregeometry

\newpage

\title{Robust Depth-based Person Re-identification}
%
%
%

\author{{Ancong Wu,
        Wei-Shi Zheng,
        Jianhuang Lai}
         \thanks{This work was supported partially by NSFC (No.61522115, 61472456, 61573387, 61661130157, 61628212), Guangdong Natural Science Funds for
Distinguished Young Scholar under Grant S2013050014265, the GuangDong Program (No.2015B010105005),
the Guangdong Science and Technology Planning Project (No.2016A010102012, 2014B010118003),
and Guangdong Program for Support of Top-notch Young Professionals (No.2014TQ01X779). The associate editor coordinating the review of
this manuscript and approving it for publication was Prof. Wen Gao.
(Corresponding author: Wei-Shi Zheng)}
        \thanks{Ancong Wu is with the School of Electronics and Information Technology, Sun Yat-sen University, Guangzhou, China; and is also with the Collaborative
Innovation Center of High Performance Computing, National University
of Defense Technology, Changsha 410073, China. Email: wuancong@mail2.sysu.edu.cn.}
        \thanks{Wei-Shi Zheng is with the School of Data and Computer Science, Sun Yat-sen University, Guangzhou, China; and is also with the Key
Laboratory of Machine Intelligence and Advanced Computing (Sun Yat-sen
University), Ministry of Education, China. E-mail: wszheng@ieee.org.}
       \thanks{Jianhuang Lai is with the School of Data and Computer Science, Sun Yat-sen University, Guangzhou, China; and is also with
Guangdong Province Key Laboratory of Information Security, P. R. China.
E-mail: stsljh@mail.sysu.edu.cn.}}

\markboth{IEEE TRANSACTIONS ON IMAGE PROCESSING}%
{Shell \MakeLowercase{\textit{et al.}}: Bare Demo of IEEEtran.cls for Journals}

\maketitle

\begin{abstract}
Person re-identification (re-id) aims to match people across non-overlapping camera views. So far the RGB-based appearance is widely used in most existing works. However, when people appeared in extreme illumination or \modifyF{changed clothes}, the RGB appearance-based re-id methods tended to fail. To overcome this problem, we propose to exploit depth information to provide more invariant body shape and skeleton information regardless of illumination and color change. More specifically, we exploit depth voxel covariance descriptor and further propose a locally rotation invariant depth shape descriptor called Eigen-depth feature to describe pedestrian body shape. We prove that the distance between any two covariance matrices on the Riemannian manifold is equivalent to the Euclidean distance between the corresponding Eigen-depth features. Furthermore, we propose a kernelized implicit feature transfer scheme to estimate Eigen-depth feature implicitly from RGB image when depth information is not available. We find that combining the estimated depth features with RGB-based appearance features can sometimes help to better reduce visual ambiguities of appearance features caused by illumination and similar clothes. The effectiveness of our models was validated on publicly available depth pedestrian datasets as compared to related methods for person re-identification.
\end{abstract}


%
\IEEEpeerreviewmaketitle

\section{Introduction}
\begin{figure}
\center
\subfigure[Clothing change]{
   \includegraphics[width=0.85\linewidth,height=0.35\linewidth]{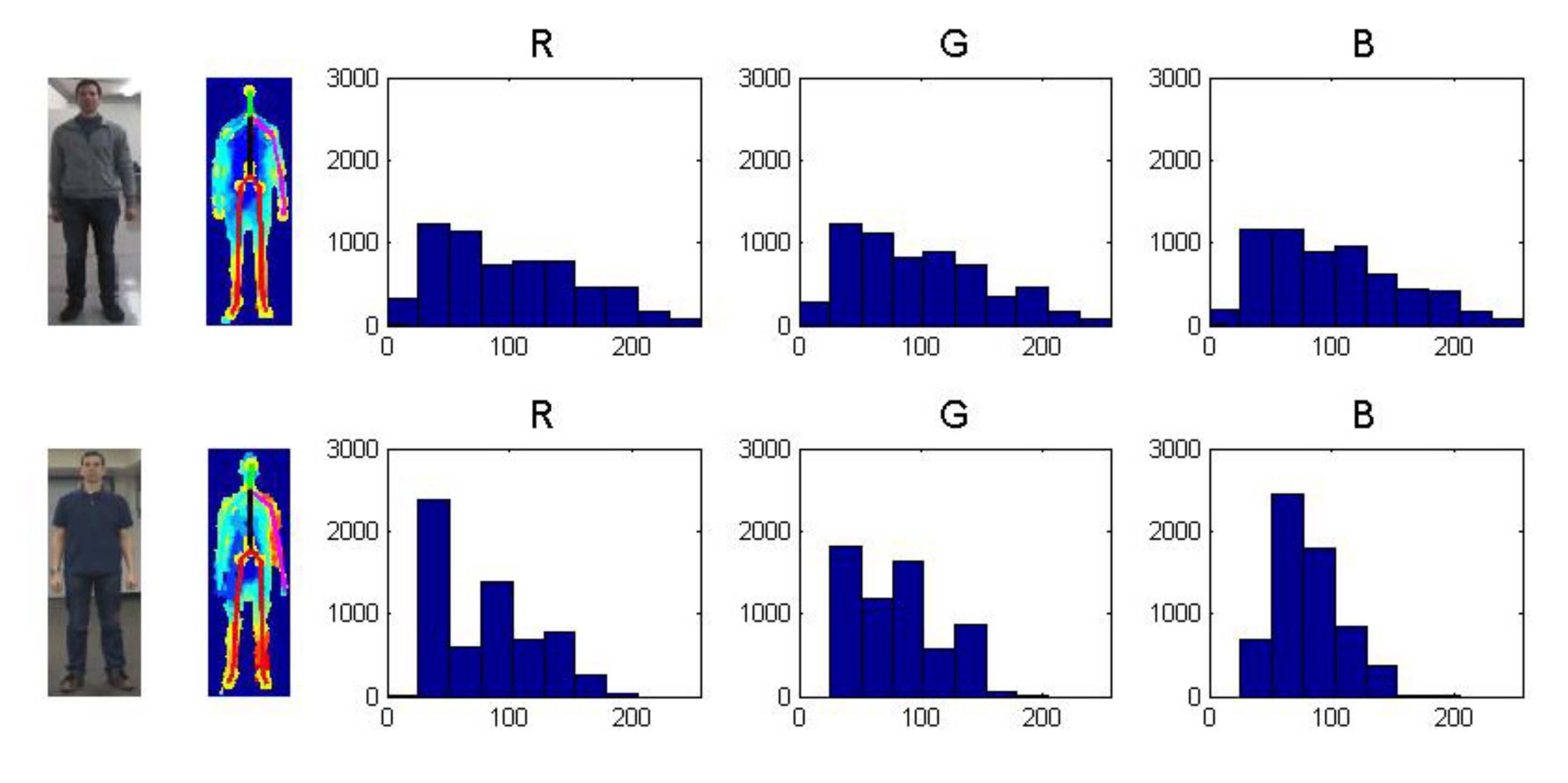}
}
\subfigure[Illumination change]{
   \includegraphics[width=0.85\linewidth,height=0.35\linewidth]{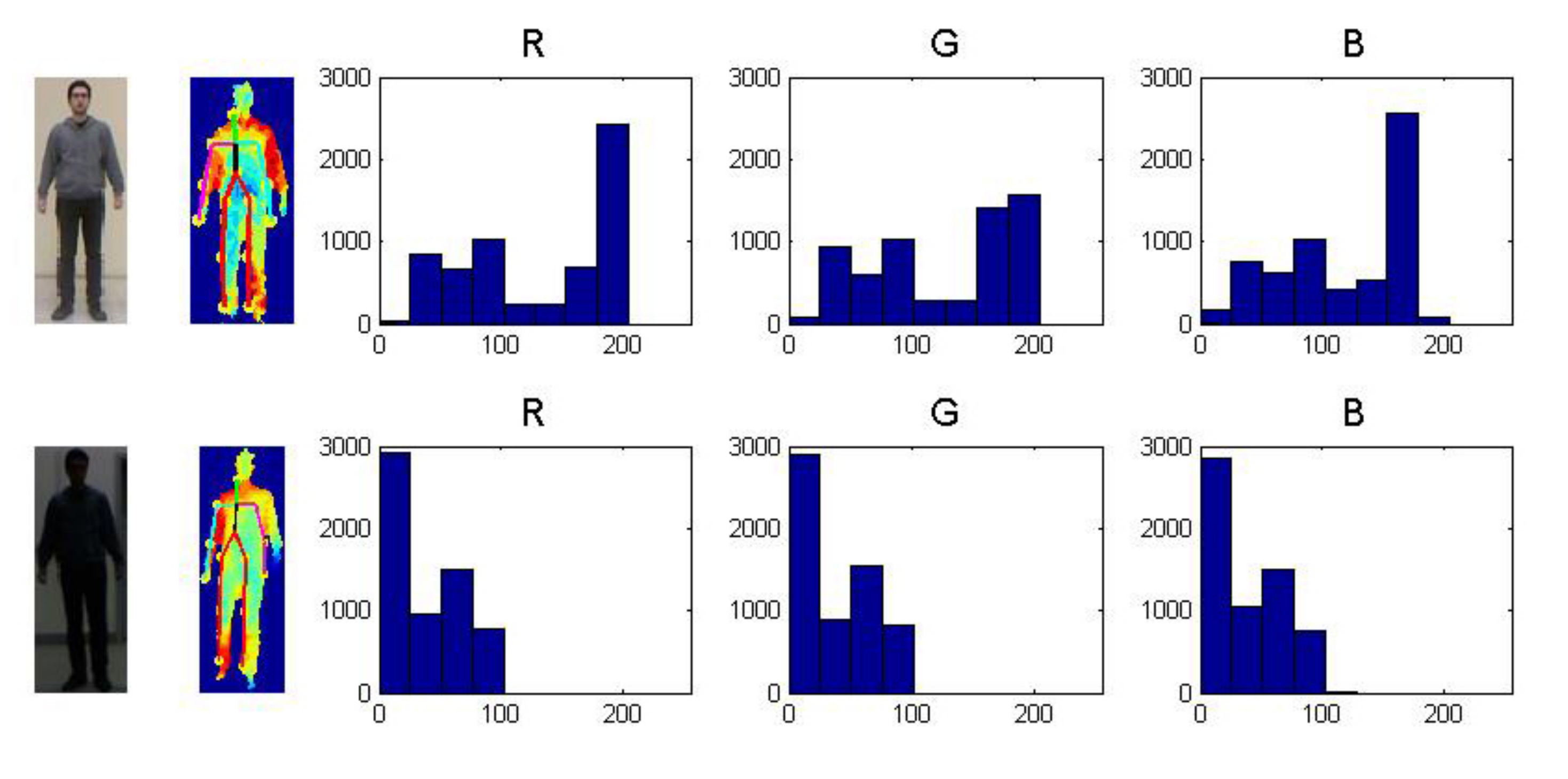}
}
\caption{Illustration of change of color histograms and invariance of depth and skeletons. From left to right, the first column shows RGB images, the second column shows depth images (shown by pseudo-color images) and skeletons, and the remaining columns show histograms of R, G, B channels, respectively.}
\label{fig:colorHistogram}
\end{figure}

The task of person re-identification \modifyD{(re-id)} is to match people in a distributed multi-camera surveillance system at different time and locations, with wide applications to forensic search, multi-camera tracking and access control, etc. In most short-term applications, low-level features such as color and textures are important appearance cues used to match. It is apparent that lighting will significantly affect the performance of these low-level features. In more extreme cases, when lighting condition changes greatly (e.g., with v.s. without lighting), color information of clothes becomes unreachable. Moreover, when people change clothes, color and textures become unreliable. For example, Figure \ref{fig:colorHistogram} shows how color histograms change when people change clothes or appear in extreme illumination. In these cases, most existing re-id systems are not workable, since they are RGB-based.

In comparison to RGB information, depth information can maintain more invariant even when suffering from clothing change and extreme illumination. As shown in Figure \ref{fig:colorHistogram}, shape and skeleton of body are likely more invariant under extreme lighting and clothing change. Nowadays, extracting depth and skeleton information with depth cameras (e.g., Microsoft Kinect) is not difficult in an indoor environment. Kinect sensor obtains depth value (distance to the camera) of each pixel by infrared, regardless of object color and illumination in indoor applications. With depth information, the life-size point cloud and skeleton of a person can be extracted, providing shape and physical information of his/her body. Moreover, with depth value of each pixel, pedestrians can be \modifyD{more} easily segmented from background, so that background influence can be largely eliminated. Hence, \modifyD{using depth information} could overcome some difficulties in RGB appearance-based methods, such as color change, illumination change and background clutters.

Although there are some advantages for depth-based re-identification as compared to the RGB appearance-based methods, challenges and limitations also come along with depth information.
\modifyB{Firstly, the depth images
captured by depth device change significantly when a person's viewpoint changes.}
Secondly, noises from devices exist in the captured depth images. \modifyB{These two aspects will seriously affect the use of depth information for person re-identification.}
So far, a few methods \cite{Barbosa:reid12,Munaro:reid13,Munaro:reid14,feifei2016CVPR} have been developed to exploit depth information for person re-identification, but the above two problems are still not well solved in existing methods. \cite{Barbosa:reid12} uses only skeletons to extract feature. In \cite{Munaro:reid13,Munaro:reid14}, besides using skeleton to extract physical information, applying point clouds converted from depth images for 3D body shape matching is also considered, but alignment errors and noises of point clouds are the problems remained unsolved. \modifyD{In \cite{feifei2016CVPR}, a deep model is applied to classify the person point cloud sequences, in which feature extraction and classification are jointly modeled and body shape is not explicitly described.} Therefore, body shape description is still an important biometric cue \modifyD{which needs further study} for person re-identification.

\begin{figure}
\center
\includegraphics[width=1\linewidth]{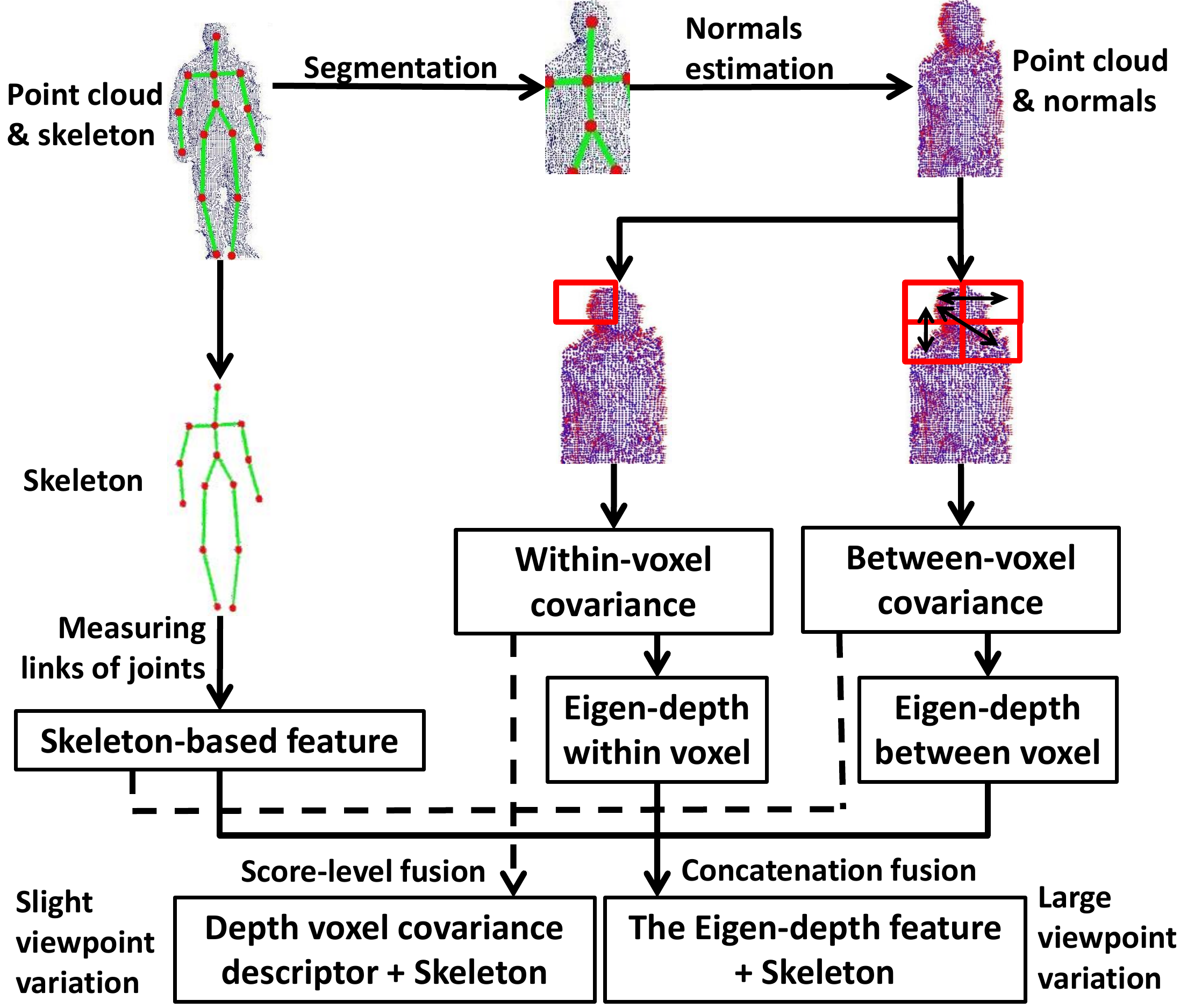}
   \caption{Pipeline of our depth-based re-identification feature extraction. In the last step, depth shape descriptor and skeleton-based feature are combined. \modifyF{When the viewpoint variation of pedestrians is slight, depth voxel covariance descriptor (DVCov) is exploited as depth shape descriptor; when the viewpoint variation is large, Eigen-depth (ED) is exploited.}}
\label{fig:pipeline}
\end{figure}

In this work, we aim to design a depth shape descriptor which is locally invariant to rotation\footnote{\modifyD{Local rotation invariance means that the feature of a body part is invariant when viewpoint change of pedestrian will not make that body part become invisible due to self-occlusion.}} and insensitive to noise. \modifyD{We propose two depth shape descriptors: depth voxel covariance descriptor and Eigen-depth feature. Eigen-depth feature is based on depth voxel covariance descriptor and locally rotation invariant.} Then we combine depth shape descriptor with skeleton-based feature to form complete depth representation of body shape and physical information. The pipeline of constructing a descriptor for our depth-based person re-id method is illustrated in Figure~\ref{fig:pipeline}. Our method takes the following steps: (1) segmentation and computation of point cloud and normals of torso and head; (2) extracting depth voxel covariance descriptor and locally rotation invariant Eigen-depth feature; (3) enriching body depth shape descriptor by additionally combining skeleton-based feature. In the second step, the Eigen-depth feature is more suitable due to its stability against local rotation of \modifyD{body} when the viewpoint of a person varies obviously,  while the depth voxel covariance descriptor will be more effective because of rich information it contains when the viewpoint change of a person is slight. \modifyC{In the third step, the skeleton-based feature can be complementary to the depth shape descriptor extracted from step (2), so more robust matching can be achieved.}

In addition, in real-world applications, most of the deployed cameras in existing surveillance systems cannot capture depth information, so how to make depth-based method work in existing system is also a challenge, \modifyD{while existing depth-based and RGB-D-based methods assume depth information is available}. Towards overcoming this limitation, we learn the relation between depth features and RGB-based appearance features by a kernelized implicit feature transfer scheme. \modifyD{For this purpose,} an auxiliary RGB-D dataset is employed to learn the nonlinear transformation between RGB-based appearance feature and depth feature. When depth device is not ready/available, the depth feature is estimated from RGB image and used to augment the RGB-based appearance feature. The experiment results show that this makes extra improvement on the re-identification performance for top-ranked matching.

We tested our methods on three publicly available datasets, PAVIS \cite{Barbosa:reid12}, BIWI RGBD-ID \cite{Munaro:reid13} and IAS-Lab RGBD-ID \cite{Munaro:reid14}. The results show the effectiveness of our depth-based approach for overcoming change of clothes and extreme illumination condition. When clothes are completely different between gallery and probe, RGB appearance-based methods fail while our depth-based method is effective. Our approach outperforms other existing depth-based re-identification methods including skeleton-based methods, PCM, combination of them \cite{Munaro:reid14} and recurrent attention model \cite{feifei2016CVPR}. Compared to other favorable rotation invariant depth shape descriptors, our descriptor also outperforms RIFT2M \cite{skelly2007improved} and Fehr's descriptor \cite{fehr2012compact}.

In summary, the contributions of our work are: (1) proposing \modifyE{depth voxel covariance descriptor and }Eigen-depth feature for depth-based re-identification and proving the local rotation invariance of Eigen-depth feature in theory; (2) forming a depth re-id recognition framework by unifying \modifyE{depth shape descriptor} and skeleton-based feature for a complete representation; (3) proposing a kernelized implicit feature transfer scheme to estimate the Eigen-depth feature from RGB images implicitly when depth device is not available.

\section{Related Work}

In this section, we present an overview of related \modifyD{image-based} re-id technologies in three aspects: (1) RGB appearance-based re-id, (2) depth-based re-id, and (3) RGB-D re-id. Currently, most person re-identification approaches are based on 2D RGB-based appearance features.


\subsection{RGB Appearance-based Person Re-identification}
\modifyB{Most existing works rely on RGB-based appearance features. \modifyD{Among them}, color is most frequently used and it is often encoded into histograms \cite{park2006vise,Farenzena10,liu2012person,kviatkovsky2013color,gray2008viewpoint,liao2015person}. Besides, texture-based features are also employed, including HOG-like signature \cite{Oreifej10}, Gabor feature \cite{gray2008viewpoint,Prosser2010person}, graph model \cite{schwartz2009learning}, differential filters \cite{gray2008viewpoint,Prosser2010person} and Haar-like representations \cite{Bak10}.
Many other hand-crafted features such as covariance descriptor \cite{B2014}, Fisher vector \cite{ma2012local}, spatial co-occurrence representation \cite{wang2007shape}, custom pictorial structure \cite{Dong2011Custom} and SARC3D \cite{baltieri20113dpes} were also developed for achieving more reliable representations.
Recently, feature learning methods have been more focused on, such as salience learning \cite{zhao2013person}, mirror representation \cite{chen2015mirror}, salient color names \cite{yang2014salient}, reference descriptor \cite{anperson2015}, \modifyF{context-based feature} \cite{wang2016zeroshot}, \modifyG{deep features} \cite{Zhao_MidLevel_2014a,Li_DeepReID_2014b,ahmed2015improved,shangxuan2016wacv,cvpr2016dropout,VariorECCV2016}, \modifyF{dictionary learning} \cite{JingCVPR2015Super,SICCV2015Person,an2016sparse} and \modifyG{attribute learning} \cite{ShiCVPR2015Transferring,ChiICCV2015Multi,SuECCV2016}. However, in the situations of clothing change or extreme illumination, these RGB-based appearance features tend to fail. 

Besides feature representation, a large amount of \modifyF{metric/subspace models} \cite{weinberger2005distance,gray2008viewpoint,kostinger2012large,Prosser2010person,zheng2013RDC,mignon2012pcca,pedagadi2013local,
li2013learning,xiong2014person,ma2014person,lisanti2014person,paisitkriangkrai2015learning,liao2015person,ChenCVPR2015Similarity,chen2015asymmetric,chenPAMI2017,LiaoICCV2015Efficient,zheng2016pami,an2016rcca,youCVPR2016,zhangCVPR2016}, have been developed to achieve more reliable matching, such as LMNN \cite{weinberger2005distance}, RankSVM \cite{Prosser2010person}, RDC \cite{zheng2013RDC}, PCCA \cite{mignon2012pcca}, KISSME \cite{kostinger2012large}, LFDA \cite{pedagadi2013local}, CVDCA \cite{chen2015asymmetric}, CRAFT \cite{chenPAMI2017}, MLAPG \cite{LiaoICCV2015Efficient}, \modifyG{TDL} \cite{youCVPR2016} \modifyG{and DNS} \cite{zhangCVPR2016}. Some other methods have also been proposed for this purpose, e.g., \modifyG{re-ranking} \cite{GarciaICCV2015Person,WangECCV2016} and correspondence structure \cite{shenICCV2015person}. \modifyG{Unsupervised learning models} \cite{pengCVPR2016,KodirovECCV2016} have also been developed for person re-identification. However, they cannot solve the illumination and clothing change problems. Compared to RGB-based appearance features, depth information is a solution to this problem, because it is independent of color and maintains more invariant for a longer period of time.}

\subsection{Depth-based Person Re-identification}

So far, only a few depth-based re-identification methods based on depth image, point cloud and anthropometric measurement \cite{Barbosa:reid12,Munaro:reid14,Munaro:reid13,Munaro:reid14b,lorenzo2012study,castrillon2014people,feifei2016CVPR}
have been developed. To some extent, depth-based methods can solve the problems of changing clothes and extreme illumination.
Barbosa et al. exploited skeleton-based feature \cite{Barbosa:reid12} based on anthropometric measurement of distances between joints and geodesic distances on body surface.
Munaro et al. built a point cloud model for each person as gallery by fusing a set of point clouds from different views and then applied Point Cloud Matching (PCM) to compute the distance between samples~\cite{Munaro:reid13}. In \cite{Munaro:reid14,Munaro:reid14b}, Munaro et al. combined PCM and skeleton-based feature modified based on Barbosa et al.'s work~\cite{Barbosa:reid12}.
These methods needed to align the point clouds, and no depth shape descriptor was applied for describing body shape. Haque et al. proposed a recurrent attention model \cite{feifei2016CVPR} for depth-video-based person identification, in which 3D RAM model was for still 3D point clouds and 4D RAM model was for 3D point cloud sequences. However, among the above depth-based frameworks, PCM and Haque's method were not suitable for solving person re-identification problem under the setting when there is no overlap on people between training and testing.

Compared to existing depth-based re-identification frameworks, the main difference of our work is that we propose \modifyE{depth voxel covariance descriptor and }Eigen-depth feature to describe body shape. Eigen-depth feature is a covariance-based feature, and it is locally rotation invariant and does not require alignment of point clouds. The Eigen-depth feature can be viewed as a depth shape descriptor and thus can remove the ambiguity of using only anthropometric measurement of skeletons in the previous depth modeling for re-identification. Compared to direct utilization of point cloud in PCM~\cite{Munaro:reid13}, it deals with noises of non-rigid human body better.

We also discuss some related depth shape descriptors, including the covariance descriptor in \cite{cov3d}, RIFT2M \cite{skelly2007improved} and Fehr's covariance descriptor \cite{fehr2012compact}, which were not applied for person re-identification.
Compared to the covariance descriptor in \cite{cov3d}, Eigen-depth feature is locally rotation invariant. Compared to rotation invariant descriptors RIFT2M \cite{skelly2007improved} and Fehr's covariance descriptor \cite{fehr2012compact}, Eigen-depth feature is densely extracted rather than using interest points, so that it contains richer information of body shape. Moreover, its rotation invariance is achieved in eigen-analysis level, so alignment of point cloud is not needed and more compact representation can be obtained by eigen-analysis.



\subsection{RGB-D Person Re-identification}

Since RGB and depth information can be obtained simultaneously when using Kinect, some re-identification methods have been developed to combine depth information and RGB appearance cues in order to extract more discriminative feature representation.
Pala et al. \cite{Pala2015rgbd} improved accuracy of clothing appearance descriptors by fusing them with anthropometric measures extracted from depth data.
Mogelmose et al. \cite{mogelmose2013tri} presented a tri-modal method to combine RGB, depth and thermal features.
Mogelmose et al. \cite{mogelmose2013multimodal} combined color histogram and height feature extracted from depth information.
John et al. \cite{john2013person} combined RGB-Height histogram and gait feature of depth information.
Satta et al. \cite{satta2013real} exploited skeleton to segment human body and extracted color feature.
In \cite{baltieri2013learning}, each color pixel was assigned to the nearest bone in the skeleton, and color histograms were computed for each region.
In \cite{albiol2012different}, the proposed feature bodyprint exploited the mean RGB values of regions in different heights.
In \cite{oliver20123d}, the descriptor was based on a 3D cylindrical grid that unified color variations together with angle and height.
Takac et al. \cite{takac2014people} exploited color histograms of upper body and lower body separately.
Xu et al. \cite{xu2015distance} proposed a distance metric using RGB-D data to improve RGB-based person re-identification. 

As reported in these works, the combination of RGB and depth is effective. They all assume that depth information is available along with RGB images. 
In our work, we propose to learn the relation between RGB and depth by a kernelized implicit feature transfer scheme, which enables estimation of depth features from RGB features so as to improve the re-identification performance even though the deployed cameras are not ready for capturing depth information.

\modifyA{
A preliminary version of this work appeared in \cite{Wu:reid15}. In this work, apart from providing more in-depth discussion on the proposed Eigen-depth feature and the depth-based person re-identification framework, a kernelized implicit feature transfer scheme is proposed to learn the relation between depth features and RGB features so as to estimate depth features in RGB images when depth sensor is not ready. In addition, more extensive experiments have been conducted.
}

\section{Depth Voxel Covariance and Eigen-depth Feature}
\label{sec:descriptor}
This section will present the extraction of depth voxel covariance descriptor and locally rotation invariant Eigen-depth feature. \modifyA{Our descriptors are extracted from point cloud, a set of points on object surface expressed by 3D coordinate $(x,y,z)$ in real world converted from depth image. \modifyF{We first tabulate the notations defined in this section in Table \ref{tab:notation1}}.

\begin{table}
\modifyF{
  \centering
  \caption{Terms and Definitions for Section \ref{sec:descriptor}.}
    \begin{tabular}{|c|c|c|c|}
    \hline
    {  symbol  } & definition & {  symbol  } & definition\\
    \hline
    $\mathbf{f}_{p,i}$ & \tabincell{c}{feature vector of the \\ $i^{th}$ point in voxel $p$ }
    &
    ${\bm \mu}_p$ & \tabincell{c}{mean feature vector of\\ voxel $p$}
    \\
    \hline
    $\mathbf{C}_{Wp}$ & \tabincell{c}{within-voxel covariance \\of voxel $p$}
    &
    $\mathbf{C}_{Bp,q}$ & \tabincell{c}{between-voxel covariance \\of voxel $p$ and $q$}
    \\
    \hline
    $\mathbf{C}_1, \mathbf{C}_2$ & \tabincell{c}{any two covariance\\ matrices}
    &
    $\mathbf{C}_{2}^N$ & \tabincell{c}{rotation normalized\\ covariance matrix from\\ $\mathbf{C}_2$ to $\mathbf{C}_1$}
    \\
    \hline
    $\mathbf{U}_{p}$ & \tabincell{c}{eigenvector matrix\\ of covariance\\ matrix $\mathbf{C}_p$}
    &
    $\lambda_{p,q}$ & \tabincell{c}{the $q^{th}$ largest\\ eigenvalue of covariance\\ matrix $\mathbf{C}_p$}
    \\
    \hline
    \end{tabular}%
  \label{tab:notation1}%
  }
\end{table}%
}

\subsection{Basic Feature Extraction}
\begin{figure}
\center
\subfigure[]{
   \includegraphics[height=0.2\linewidth]{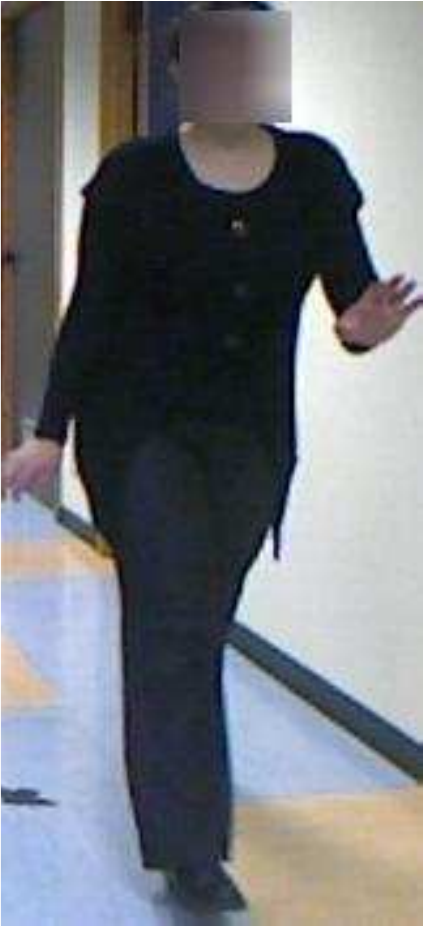}
}
\subfigure[]{
   \includegraphics[height=0.2\linewidth]{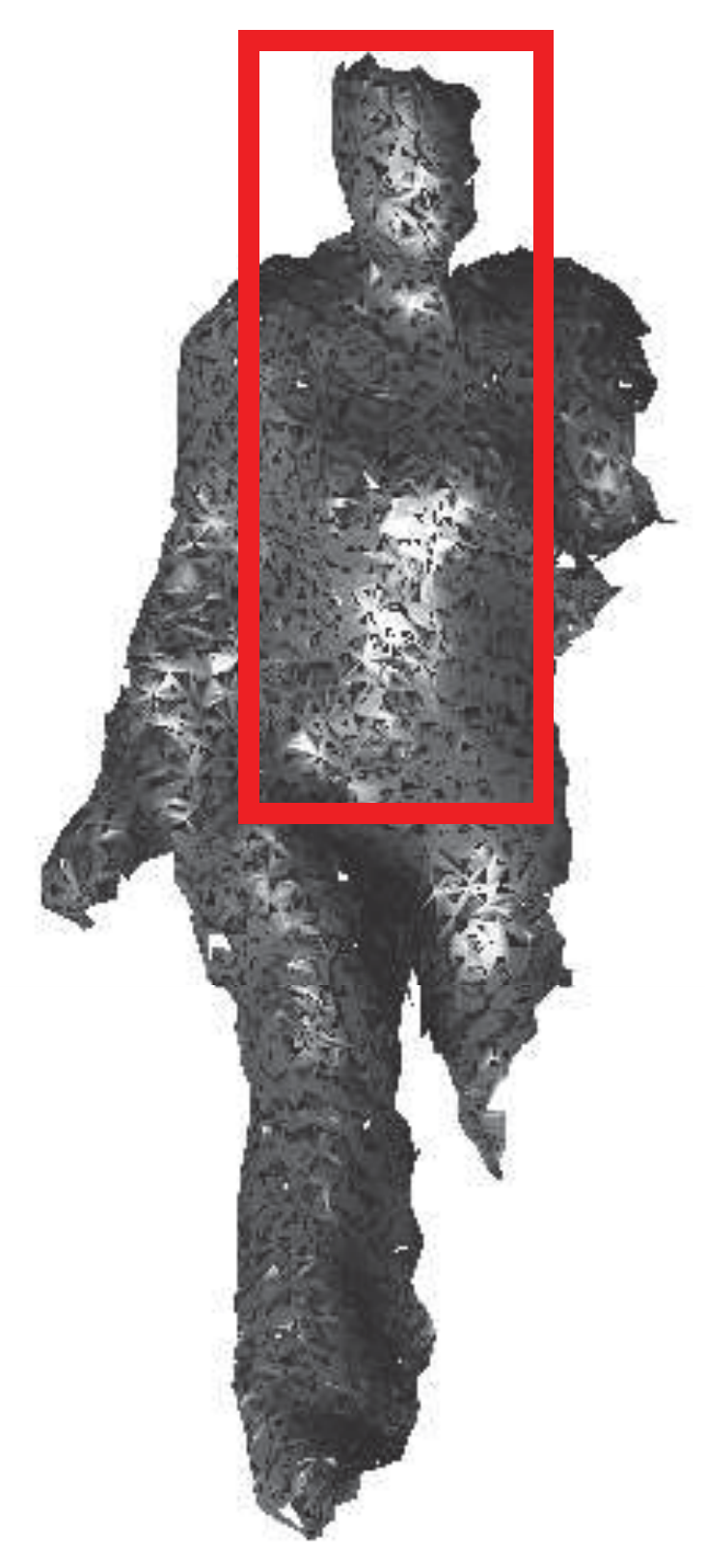}
}
\subfigure[]{
   \includegraphics[height=0.2\linewidth]{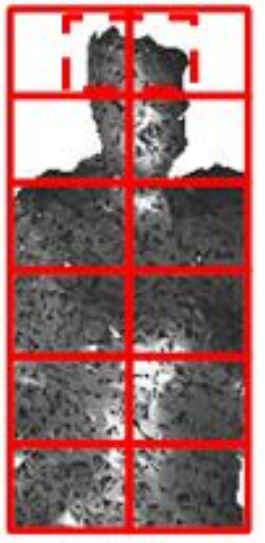}
}
\subfigure[]{
   \includegraphics[height=0.2\linewidth]{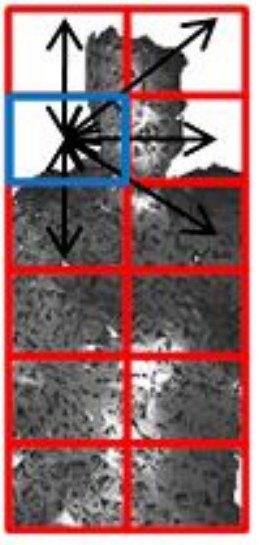}
}
\caption{Illustration of body self-occlusion and feature extraction region. (a) is a sample RGB image, (b) is the corresponding point cloud, (c) is the within-voxel feature extraction region and (d) is the between-voxel feature extraction region.}
\label{fig:mesh}
\end{figure}

We first extract basic features of point cloud. We assume another kind of biometric cue, skeleton joints of pedestrian body, is also available along with depth images (e.g., when using Kinect). 
We intend to extract features on the body parts whose surfaces are more invariant and reliable. As shown in Figure~\ref{fig:mesh} (a) and (b), due to pose difference, sometimes a part of limb surface is not observed under self-occlusion, so the surface shapes of arms and legs contain more noises rather than valuable information. Therefore, we divide each point cloud of the whole body by two shoulder joints and two hip joints and only the points of head and torso are used for feature extraction, while the four limbs are not.

For each point in the point cloud, a normal vector \cite{hoppe1992surface} is computed as basic feature. The direction of normal vector describes the shape of a small neighbour region of that point. For a point $\mathbf{x}$, $k$ nearest neighbourhoods of $\mathbf{x}$ are found, \modifyA{and then the direction on which data is least scattered is computed by PCA \cite{lda}} as the unit normal vector direction $(n_x,n_y,n_z)$. For each point $(x,y,z)$ (unit: mm), a feature vector $F(x,y,z)$ is composed of the coordinate and the unit normal vector
\begin{equation}
F(x,y,z)=[x,y,z,n_x,n_y,n_z]^T.
\end{equation}

\subsection{Depth Voxel Covariance Descriptor}
\label{section:loal_depth}
To depict the variation of local feature vectors and alleviate noises, we exploit two types of covariance matrices, namely within-voxel covariance and between-voxel covariance.

\subsubsection{Within-voxel Covariance}
\label{within}
We divide a point cloud into rectangular \modifyA{voxels} (e.g., $6\times 2$ voxels in our case) with 50\% overlap, and an example is shown in Figure~\ref{fig:mesh} (c). In each voxel, within-voxel covariance matrix is computed to describe the shape. For a voxel $R_1$, let $\{\mathbf{f}_{1,i}\}_{i=1}^m$ be the 6-dimensional feature vectors inside $R_1$. Within-voxel covariance matrix \modifyF{$\mathbf{C}_{W1}$} is then defined as follows:
\modifyF{
\begin{equation}
\mathbf{C}_{W1}=\frac{1}{m-1}\sum_{i=1}^{m}(\mathbf{f}_{1,i}-\bm{\mu}_1)(\mathbf{f}_{1,i}-\bm{\mu}_1)^T,
 \label{equ:3}
\end{equation}
}
where $\bm{\mu}_1$ is the mean of the feature vectors of $R_1$.

\subsubsection{Between-voxel Covariance}
\label{between}
While within-voxel covariance describes shape in a voxel, the differences of shapes between voxels also contain discriminative information. Similar to standard covariance, we define a novel between-voxel covariance to represent the relation between different voxels.

As shown in Figure~\ref{fig:mesh} (d), the point cloud is divided into $6\times 2$ voxels without overlap. Between-voxel covariance matrices are computed for each pair of 8-adjacent voxels. For two adjacent voxels $R_1$ and $R_2$, let $\{\mathbf{f}_{1,i}\}_{i=1}^m$ and $\{\mathbf{f}_{2,j}\}_{j=1}^n$ be the 6-dimensional feature vectors inside $R_1$ and $R_2$, respectively. We define the between-voxel covariance matrix \modifyF{$\mathbf{C}_{B1,2}$} as follows:
\modifyF{
\begin{equation}
\mathbf{C}_{B1,2}=\frac{1}{mn}\sum_{i=1}^{m}\sum_{j=1}^{n}(\mathbf{f}_{1,i}-\mathbf{f}_{2,j})(\mathbf{f}_{1,i}-\mathbf{f}_{2,j})^T.
\label{equ:4}
\end{equation}
}

For a depth image, the unification of the within-voxel and between-voxel covariance matrices of all voxels is called the \emph{depth voxel covariance descriptor} (DVCov).

\subsection{\modifyE{Local Rotation Invariance of Eigenvalues}}
\label{eigenvalue}

Assume $p$, $q$ are voxels of a person in depth camera $K$. Let $\{\mathbf{f}^K_{p,i}\}_{i=1}^N$ denote the feature vectors, \modifyF{$\mathbf{C}^{K}_{Wp}$} denote the within-voxel covariance matrix of voxel $p$ and \modifyF{$\mathbf{C}^{K}_{Bp,q}$} denote between-voxel covariance matrix between voxels $p$ and $q$. We assume only viewpoint rotation and location change take place between two different camera views $A$ and $B$ \modifyE{and the rotation is local so that the body part within voxels $p$ and $q$ will not become invisible due to self-occlusions}. To express the transformation from camera view $A$ to camera view $B$, let $\mathbf{R}_{AB1}$ denote the rotation transformation matrix of point coordinate $(x,y,z)$, $\mathbf{R}_{AB2}$ denote the rotation transformation matrix of unit normal vector $(n_x,n_y,n_z)$, and $\mathbf{f}_{AB}=[x_s,y_s,z_s,0,0,0]^T$ denote the shift of pedestrian location. Then the transformations of feature vectors from camera view $A$ to camera view $B$ are
 \begin{equation}
 \mathbf{f}^B_{1,i}= \mathbf{R}_{AB} (\mathbf{f}^A_{1,i}+ \mathbf{f}_{AB}),
  \label{equ:6}
 \end{equation}
 \begin{equation}
 \mathbf{f}^B_{2,i}= \mathbf{R}_{AB} (\mathbf{f}^A_{2,i}+ \mathbf{f}_{AB}),
  \label{equ:7}
 \end{equation}
where $\mathbf{R}_{AB}=
  \left(
  \begin{array}{cc}
    \mathbf{R}_{AB1} & \mathbf{O}\\
    \mathbf{O} & \mathbf{R}_{AB2}\\
  \end{array}
  \right)
  $.

By substituting equations (\ref{equ:6}), (\ref{equ:7}) into (\ref{equ:3}), (\ref{equ:4}), we have
\modifyF{
 \begin{equation}
 \mathbf{C}^B_{W1}=\mathbf{R}_{AB} \mathbf{C}^A_{W1} \mathbf{R}_{AB}^T, \mathbf{C}^B_{B1,2}=\mathbf{R}_{AB} \mathbf{C}^A_{B1,2} \mathbf{R}_{AB}^T.
 \label{equ:8}
 \end{equation}
}

Since $\mathbf{R}_{AB1}$ and $\mathbf{R}_{AB2}$ satisfy $\mathbf{R}_{AB1}\mathbf{R}_{AB1}^T=\mathbf{I}$ and $\mathbf{R}_{AB2}\mathbf{R}_{AB2}^T=\mathbf{I}$, we have $\mathbf{R}_{AB} \mathbf{R}_{AB}^T=\mathbf{I}$, so that $\mathbf{R}_{AB}$ is  orthogonal transformation. Hence, the eigenvalues of within-voxel covariance matrices \modifyF{$\mathbf{C}^B_{W1}$} and \modifyF{$\mathbf{C}^A_{W1}$} are the same, and the eigenvalues of between-voxel covariance matrices \modifyF{$\mathbf{C}^B_{B1,2}$ and $\mathbf{C}^A_{B1,2}$} are the same as well.
That means the eigenvalues of within-voxel covariance and between-voxel covariance are invariant to rotation and location change.

\subsection{\modifyE{Eigen-depth Feature and Analysis}}\label{eigen-depth}

In this section, we provide more in-depth analysis about the role of those eigenvalues. Let $\mathbf{C}_1,\mathbf{C}_2\in Sym^+(6,\mathbb{R})$ denote two covariance matrices. 
The eigen-decomposition of $\mathbf{C}_1$ and $\mathbf{C}_2$ are $\mathbf{C}_1=\mathbf{U}_1 {\rm diag}(\lambda_{1,1},\lambda_{1,2},...,\lambda_{1,6}) \mathbf{U}_1^T$ and $\mathbf{C}_2=\mathbf{U}_2 {\rm diag}(\lambda_{2,1},\lambda_{2,2},...,\lambda_{2,6}) \mathbf{U}_2^T$, respectively.
Here $\lambda_{1,1}\geq \lambda_{1,2}\geq...\geq \lambda_{1,6}$ are eigenvalues of $\mathbf{C}_1$, $\lambda_{2,1}\geq \lambda_{2,2}...\geq\lambda_{2,6}$ are eigenvalues of $\mathbf{C}_2$, and $\mathbf{U}_1$ and $\mathbf{U}_2$ are orthogonal matrices whose columns are the corresponding eigenvectors.

We note that rotation of point clouds and normal vectors can be normalized by matching the principal axes of $\mathbf{C}_2$ and $\mathbf{C}_1$ according to the descending order of eigenvalues. That is, one can find an orthogonal transformation matrix $\mathbf{Q}$ such that $\mathbf{Q} \mathbf{U}_2 = \mathbf{U}_1$, where $\mathbf{Q}$ is the rotation transformation we want to estimate. Hence, we construct a normalized covariance matrix $\mathbf{C}_{2}^N= \mathbf{U}_1 {\rm diag}(\lambda_{2,1},\lambda_{2,2},...,\lambda_{2,6}) \mathbf{U}_1^T$, where $\lambda_{2,1},\lambda_{2,2},...,\lambda_{2,6}$ are eigenvalues of $\mathbf{C}_2$ and $\mathbf{U}_1$ contains eigenvectors of $\mathbf{C}_1$. We call $\mathbf{C}_2^N$ the rotation normalized covariance matrix from $\mathbf{C}_2$ to $\mathbf{C}_1$.

Now we present how to use the above eigenvalues to construct feature vectors. Let $\mathbf{x}_1=[{\rm ln}\lambda_{1,1}~{\rm ln}\lambda_{1,2}~...~{\rm ln}\lambda_{1,6}]^T$ and $\mathbf{x}_2=[{\rm ln}\lambda_{2,1}~{\rm ln}\lambda_{2,2}~...~{\rm ln}\lambda_{2,6}]^T$. Interestingly, we can have the following theorem.

\noindent \begin{thm}
Computing the Euclidean distance between $\mathbf{x}_1$ and $\mathbf{x}_2$ is equivalent to computing the geodesic distance between covariance matrix $\mathbf{C}_1$ and the rotation normalized covariance matrix $\mathbf{C}_2^N$ on the Riemannian manifold.
\end{thm}

\textbf{Proof.}
The Euclidean distance between $\mathbf{x}_1$ and $\mathbf{x}_2$ is
\begin{equation}
\begin{aligned}
\|\mathbf{x}_2-\mathbf{x}_1\|_2
 =\sqrt{\sum^6_{i=1}({\rm ln}\lambda_{2,i}-{\rm ln}\lambda_{1,i})^2}
 =\sqrt{\sum^6_{i=1}{\rm ln}^2 \frac{\lambda_{2,i}}{\lambda_{1,i}} }.
\end{aligned}
\label{equ:12}
\end{equation}

The geodesic distance between $\mathbf{C}_1$ and $\mathbf{C}_{2}^N$ on Riemannian manifold \cite{forstner2003metric} can be calculated as follows:
\begin{equation}
dist(\mathbf{C}_1,\mathbf{C}_{2}^N)=\sqrt{\sum^6_{k=1}{\rm ln}^2 \lambda_k(\mathbf{C}_1,\mathbf{C}_{2}^N) },
\label{equ:13}
\end{equation}
where $\lambda_k(\mathbf{C}_1,\mathbf{C}_{2}^N)_{k=1,...,6}$ are the generalized eigenvalues of $\mathbf{C}_1$ and $\mathbf{C}_{2}^N$, computed by $\lambda_k \mathbf{C}_1 \mathbf{z}_k-\mathbf{C}_{2}^N \mathbf{z}_k=0$, 
i.e., eigenvalues of $\mathbf{C}_1^{-1}\mathbf{C}_{2}^N$.
\begin{equation}
\begin{aligned}
\mathbf{C}_1^{-1}\mathbf{C}_{2}^N &=(\mathbf{U}_1 {\rm diag}(\lambda_{1,i}) \mathbf{U}_1^T)^{-1} (\mathbf{U}_1 {\rm diag}(\lambda_{2,i}) \mathbf{U}_1^T)\\
& =\mathbf{U}_1 {\rm diag}(\frac{\lambda_{2,i}}{\lambda_{1,i}}) \mathbf{U}_1^T.
\end{aligned}
\end{equation}
Hence, the $i^{th}$ generalized eigenvalue of $\mathbf{C}_1$ and $\mathbf{C}_{2}^N$ is
\begin{equation}
\lambda_i(\mathbf{C}_1,\mathbf{C}_{2}^N) = \frac{\lambda_{2,i}}{\lambda_{1,i}}.
\label{equ:16}
\end{equation}
By substituting Equation (\ref{equ:16}) into (\ref{equ:12}) and (\ref{equ:13}), we have
\begin{equation}
\|\mathbf{x}_2-\mathbf{x}_1\|_2=dist(\mathbf{C}_1,\mathbf{C}_{2}^N).
\end{equation}

It can be seen that the geodesic distance on the Riemannian manifold $dist(\mathbf{C}_1,\mathbf{C}_{2}^N)$ is equivalent to the Euclidean distance between feature vectors $\mathbf{x}_1$ and $\mathbf{x}_2$. 

\begin{figure}
\center
\includegraphics[width=1\linewidth]{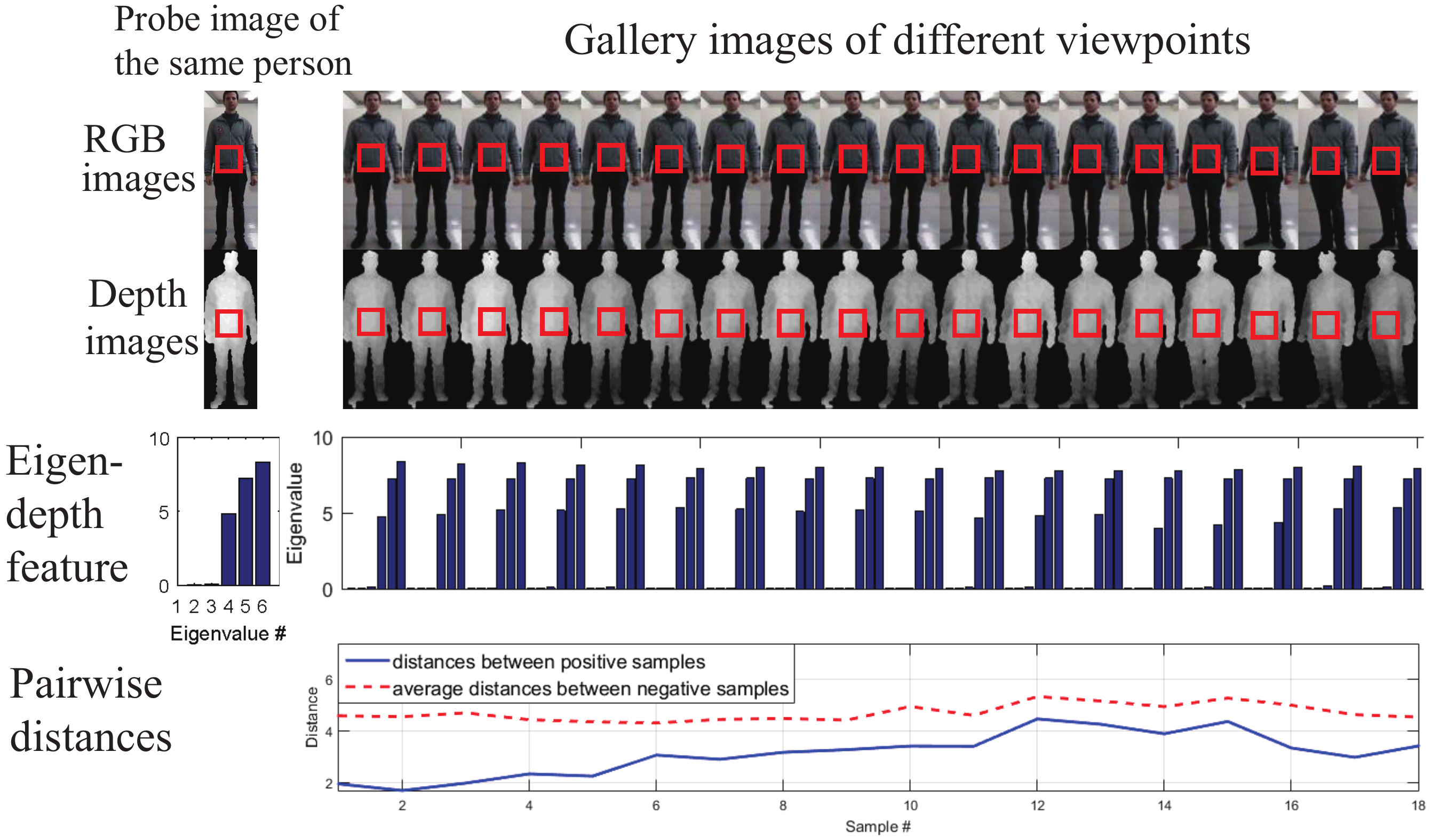}
   \caption{\modifyF{Visualization of the logarithms of eigenvalues of a fixed voxel and comparison of distance of positive pair (i.e., samples from the same class) and average distance of negative pair (i.e., samples from different classes). The first row shows RGB and depth images of the same person. The second row shows within-voxel Eigen-depth feature of the fixed voxel indicated by red bounding boxes. The third row shows the comparison between the distance of positive pair and the distance of negative pair.}}
\label{fig:eigenvalue}
\end{figure}

\noindent \textbf{Eigen-depth Feature}. The above theorem tells if there exists only local rotation variation, the logarithm eigenvalue vector can convert the distance between covariance matrices on Riemannian manifold to the Euclidean distance between two feature vectors. 
In our work, we define the Eigen-depth feature (ED) of a covariance matrix $\mathbf{C}_p$ as
\begin{equation}
\mathbf{x}_p=[{\rm ln} \lambda_{p,1}~{\rm ln} \lambda_{p,2}~...~{\rm ln} \lambda_{p,6}]^T,
\end{equation}
\modifyD{where $\mathbf{C}_p$ is either a within-voxel covariance or a between-voxel covariance.}
Using eigenvalues makes the feature more compact than using depth voxel covariance descriptor.

\modifyF{
To give a direct perception of Eigen-depth features, i.e., the logarithms of eigenvalues, we show some sample images, the Eigen-depth features and distances between positive and negative pairs in Figure \ref{fig:eigenvalue}. For demonstration, we selected one sample as probe image from BIWI RGBD-ID dataset and 18 samples of the same person captured from different views as gallery images. For a fixed voxel indicated in the red bounding box, we extracted its within-voxel Eigen-depth feature and obtained a 6-dimensional feature vector consisting of the logarithms of eigenvalues for each sample. The logarithms of eigenvalues are shown in the second row in Figure \ref{fig:eigenvalue} by the histograms. We can find that the histograms of eigenvalues look very similar over the rotation change. Since there are still extra variations but not just local rotation variation in practice, we further make comparison between the distance of positive pair (i.e., samples from the same class) and the distance of negative pair (i.e., samples from different classes) on the third row of Figure \ref{fig:eigenvalue}. We first computed the Euclidean distance between the probe image and each gallery image given above as the distance of positive pair (plotted as blue curve), and computed the average distance between each gallery image and all samples from other classes in this dataset as the distance of negative pair (plotted as red dashed curve). We can observe that the distance between samples of the same class across multiple view angles is less sensitive to rotation in practice. Moreover, the distance of positive pair is smaller than the average distance of negative pair. So the Eigen-depth feature is a useful shape descriptor for recognition.
}

\vspace{0.1cm}
\noindent \textbf{Remark.} In existing literatures about covariance descriptor such as \cite{B2014}, geodesic distance on Riemannian manifold is used for measuring the similarities between covariance matrices. However, directly using geodesic distance is not invariant to rotation. Given two rotation transformation matrices $\mathbf{R}_1$ and $\mathbf{R}_2$, the eigenvalues of $\mathbf{C}_1^{-1}\mathbf{C}_2$ and $\mathbf{R}_1\mathbf{C}_1^{-1}\mathbf{R}_1^T\mathbf{R}_2\mathbf{C}_2\mathbf{R}_2^T$ are always different. Moreover, covariance matrix does not lie on Euclidean space, so most common machine learning methods are not proper to be applied directly.

\begin{figure}
\center
\includegraphics[width=0.4\linewidth]{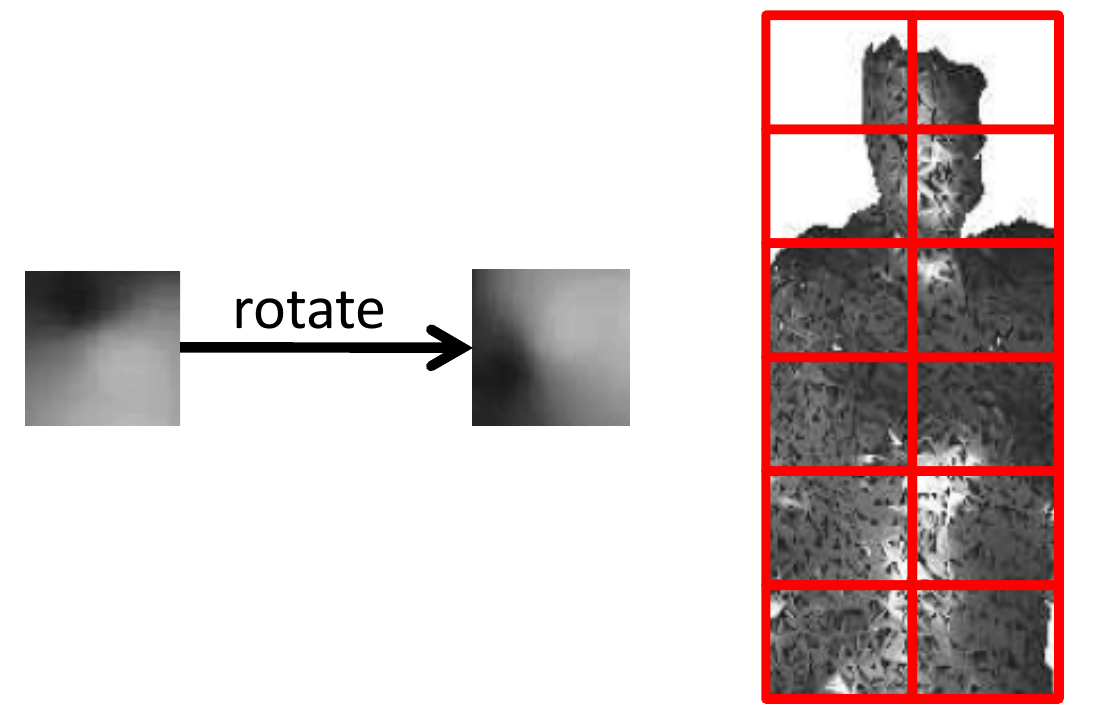}
   \caption{The left illustrates the confusion caused by rotation invariance, and the right shows how voxels are divided.}
\label{fig:patchSize}
\end{figure}

In practice, although Eigen-depth feature is proved to be locally invariant to rotation, some problems come along with this property. As shown in Figure \ref{fig:patchSize} on the left, given depth images (in which grayscale denotes depth) of two different voxels of body surface, they can be transformed to each other by rotation. Obviously, their shapes are clearly different but the Eigen-depth features of within-voxel covariance matrices are the same, and such a situation could make confusion in the matching stage, which is also a problem for other rotation invariant depth shape descriptors. This kind of confusion could take place if the voxel size is too small and the voxel contains only a small region of body surface.
To alleviate this problem, as illustrated in Figure \ref{fig:patchSize} on the right, we divide the point cloud into $6\times 2$ voxels to extract feature for more robust representation. So the voxels are large enough to contain a big area of body surface, making it less possible to cause confusion after rotation.

\section{Depth-based Re-identification Framework}

In the previous section, we have extracted depth voxel covariance descriptors (DVCov), and constructed Eigen-depth feature (ED) for describing body shape. \modifyE{Besides using body shape}, incorporating more physical information would have extra benefit on the identification of a person. As indicated in the previous section, the four limbs are not suitable for extracting invariant shape representation, but the lengths of limbs contain physical information which is also a biometric cue for distinguishing people. Hence, to obtain a complete feature representation of pedestrian, we additionally employ the skeleton-based feature (SKL) as complementary physical information, and then build a depth-based re-identification framework by combining the proposed depth shape descriptors and the skeleton-based feature together.




The whole framework is illustrated in Figure \ref{fig:pipeline}. For the feature representation of skeleton, we apply the skeleton-based feature in~\cite{Munaro:reid13}. This skeleton-based feature is a feature vector that contains 13 distance values and ratios computed from the positions of skeleton joints provided by a skeleton tracker. The elements of the feature vector includes: (a) head height, (b) neck height, (c) neck to left shoulder distance, (d) neck to right shoulder distance, (e) torso to right shoulder distance, (f) right arm length, (g) left arm length, (h) right upper leg length, (i) left upper leg length, (j) torso length, (k) right hip to left hip distance, (l) ratio between torso length and right upper leg length (i.e., j/h) and (m) ratio between torso length and left upper leg length (i.e., j/i) (the unit of distances is cm).

After extracting skeleton-based feature, in the stage of feature fusion, we combine our proposed depth shape descriptors and the skeleton-based feature together to form complete representation of human body. In this work, we offer two fusion models below.
\begin{itemize}
  \item \textbf{DVCov+SKL:} When the viewpoint variation of a person across camera views is not large in some special cases (e.g., security check or walking in narrow passage), the influence of rotation can be secondary. In such cases, we select our depth voxel covariance descriptor as depth shape descriptor, as it contains richer information about texture and is more effective for describing the shape. We measure the similarity of two subjects by computing the combined distance $d=d_{DVCov}+d_{SKL}$, where $d_{DVCov}$ denotes the sum of geodesic distances \modifyD{between the corresponding within-voxel covariance matrices and between-voxel covariance matrices}, and $d_{SKL}$ denotes the Euclidean distance between skeleton-based features.
  \item \textbf{ED+SKL:} When the viewpoint variation of a pedestrian across different camera views is large, we select Eigen-depth feature as depth shape descriptor since it is locally rotation invariant. Let $\mathbf{x}_W$ and $\mathbf{x}_B$ denote the concatenated Eigen-depth feature vectors of all within-voxel covariance matrices and all between-voxel covariance matrices of a person. Let $\mathbf{x}_{SKL}$ denote the skeleton-based feature. We fuse these three features by concatenating them to obtain a combined feature $\mathbf{x}_C=[\mathbf{x}_W^T~\mathbf{x}_B^T~\mathbf{x}_{SKL}^T]^T$. Then we apply LDA \cite{lda} to $\mathbf{x}_C$ for feature selection. We first reduce feature dimension to 100 by principal component analysis (PCA) and then extract $c-1$ discriminant vectors by LDA, where $c$ is the number of classes. After dimension reduction, the projected features are matched by using Euclidean distance.
\end{itemize}



\section{Depth Feature Transfer}
\label{sec:depth_transfer}
We have proposed a depth-based person re-identification framework in previous sections. However, in most existing surveillance systems, a large amount of cameras do not support capturing depth information, so only RGB images are available. In this section, we exploit a transfer technique to implicitly estimate depth features for RGB person images when depth device is not ready. \modifyF{We tabulate the notations defined in this section in Table \ref{tab:notation2}}.

\begin{table}
\modifyF{
  \centering
  \caption{\modifyF{Terms and Definitions for Section \ref{sec:depth_transfer}.}}
    \begin{tabular}{|c|c|c|c|}
    \hline
    {  symbol  } & definition & {  symbol  } & definition\\
    \hline
    $\mathbf{f}_{i}^v$, $\mathbf{f}_{i}^d$ & \tabincell{c}{visual/depth feature of the \\ $i^{th}$ sample }
    &
    $y_i$ & \tabincell{c}{label of the $i^{th}$ sample}
    \\
    \hline
    $\phi_v(\cdot)$, $\phi_d(\cdot)$ & \tabincell{c}{nonlinear mapping for\\ visual/depth feature}
    &
    $\mathbf{W}_v$, $\mathbf{W}_d$ & \tabincell{c}{projection matrix for\\ nonlinear visual/depth\\ feature}
    \\
    \hline
    $\mathbf{S}_{bv}$, $\mathbf{S}_{bd}$ & \tabincell{c}{between-class scatter\\ matrix for visual/depth\\ feature}
    &
    $\mathbf{S}_{wv}$, $\mathbf{S}_{wd}$ & \tabincell{c}{within-class scatter\\ matrix for visual/depth\\ feature}
    \\
    \hline
    \tabincell{c}{${\rm K}_{v}(\cdot,\cdot)$ , \\ ${\rm K}_{d}(\cdot,\cdot)$} & \tabincell{c}{kernel function for \\ visual/depth features}
    &
    $\mathbf{A}_v$, $\mathbf{A}_d$ & \tabincell{c}{combination coefficient\\ matrix for visual/depth\\ features}
    \\
    \hline
    \end{tabular}%
  \label{tab:notation2}%
  }
\end{table}%

\subsection{Kernelized Implicit Feature Transfer Scheme}
Depth features can describe body shape of a person, while some visual features (e.g., HOG \cite{Oreifej10} and LBP \cite{lbp}) extracted from RGB images can also describe body shape coarsely to some extent. Therefore, we aim to learn the relation between depth features and these kinds of visual features, so as to estimate the depth features from RGB images when depth device is not ready.

\begin{figure}
\center
\includegraphics[width=1\linewidth]{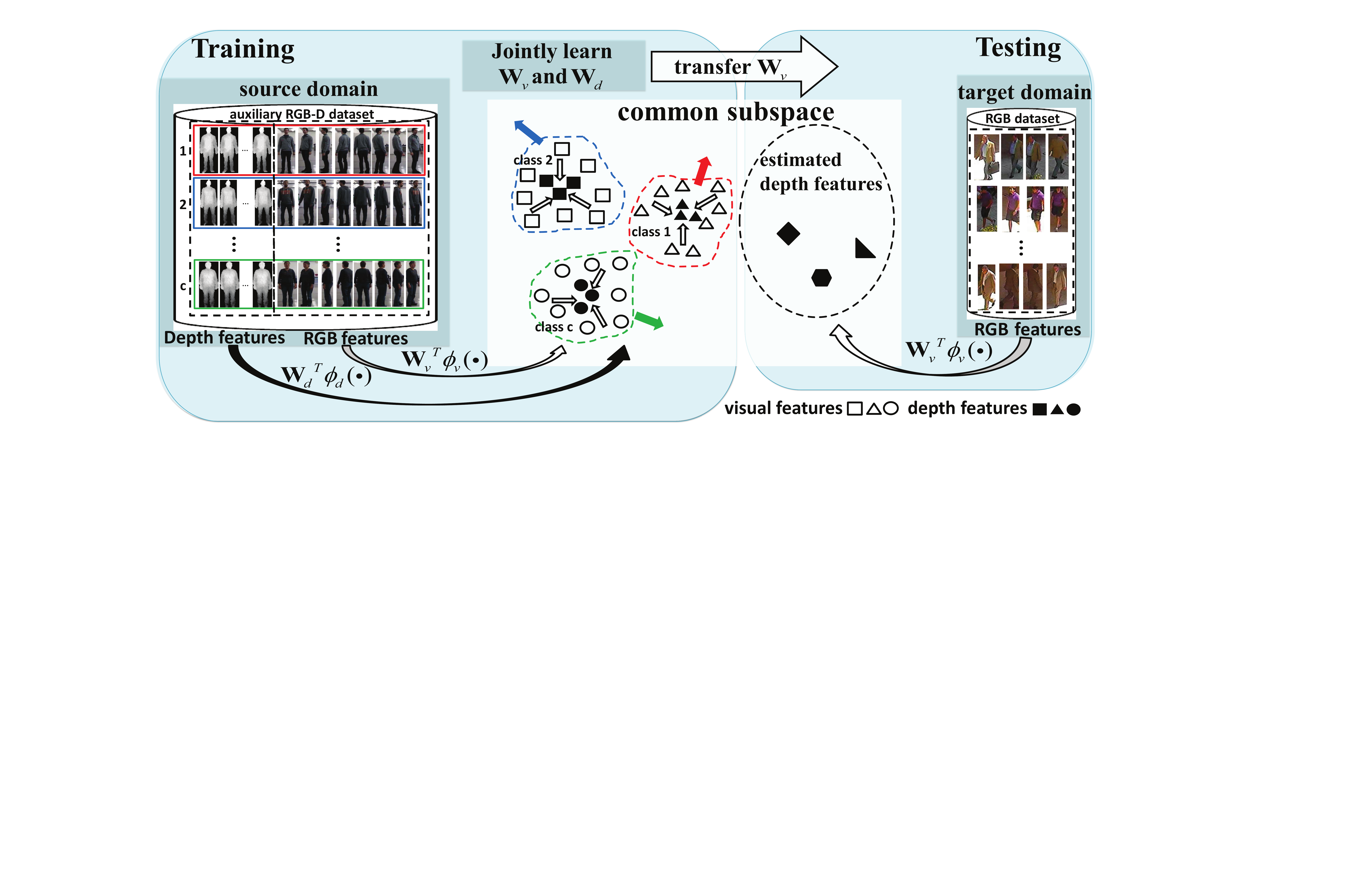}
\caption{Overview of kernelized implicit feature transfer scheme. Visual features and depth features are mapped by projections to a common feature subspace.}
\label{fig:transfer_pipeline}
\end{figure}

For this purpose, we assume an auxiliary RGB-D dataset is given. This RGB-D dataset is regarded as source domain, and the RGB images from which we want to estimate depth features are regarded as target domain. We propose a kernelized implicit feature transfer scheme to transfer the depth feature from source domain to target domain. The overview of the feature transfer procedure is shown in Figure \ref{fig:transfer_pipeline}.

In details, suppose there exists an auxiliary RGB-D dataset that consists of RGB-D images for each person. Let the source domain samples be denoted by $\{(\mathbf{f}_{i}^v,\mathbf{f}_i^d,y_i)\}_{i=1}^{N_s}$, where $N_s$ is the total number of samples, $\mathbf{f}_{i}^v$ denotes the visual feature of the $i^{th}$ sample, $\mathbf{f}_i^d$ denotes the depth feature of the $i^{th}$ sample, and \modifyD{$y_i\in\{1,2,...,C\}$} denotes the label ($C$ is the total number of classes/identities).
Depth feature and visual feature are heterogeneous features, and we assume that they can be mapped \modifyF{onto} a common latent subspace if they are transformed onto high dimensional nonlinear space implicitly by some kernel functions. Let us denote the nonlinear visual feature as $\phi_v(\mathbf{f}^v) \in \mathbb{R}^{m_v}$ and the nonlinear depth feature as $\phi_d(\mathbf{f}^d) \in \mathbb{R}^{m_d}$, where the dimensions $m_v$ and $m_d$ are unknown. Then we project the visual features and depth features onto a common latent subspace by projection matrices $\mathbf{W}_v \in \mathbb{R}^{m_v \times m}$ and $\mathbf{W}_d \in \mathbb{R}^{m_d \times m}$, respectively, where $m$ is the dimension of the common latent subspace. In the common latent subspace, we aim to make the projected visual feature $\mathbf{W}_v^T \phi_v(\mathbf{f}^v)$ close to the corresponding depth feature $\mathbf{W}_d^T \phi_d(\mathbf{f}^d)$. For this purpose,
we minimize the distance between the means of RGB-based visual features and the depth features of each person in the common latent subspace by minimizing
\begin{equation}
\Omega_{vd}= \frac{1}{C}\sum_{c=1}^{C} \norm*{\frac{1}{N_c}\sum_{y_i=c}\mathbf{W}_v^T \phi_v(\mathbf{f}_{i}^v) - \frac{1}{N_c}\sum_{y_i=c}\mathbf{W}_d^T \phi_d(\mathbf{f}_i^d)}^2_2.
\label{equ:MMD_heter}
\end{equation}


In addition, we wish that the above transformation between depth and RGB features is learned in a discriminative way.
In order to make both visual features and depth features discriminative in the common latent subspace, it is expected to minimize the within-class variance and maximize the between-class variance of both visual features and depth features at the same time.
The between-class scatter matrices and within-class scatter matrices are defined as follows:
\begin{equation}
\mathbf{S}_{b*}=\sum_{i,j=1}^{N_s} \mathbf{A}_{i,j}^b (\phi_*(\mathbf{f}_i^*)-\phi_*(\mathbf{f}_j^*))(\phi_*(\mathbf{f}_i^*)-\phi_*(\mathbf{f}_j^*))^T,
\label{equ:scatter_between}
\end{equation}
\begin{equation}
\mathbf{S}_{w*}=\sum_{i,j=1}^{N_s} \mathbf{A}_{i,j}^w (\phi_*(\mathbf{f}_i^*)-\phi_*(\mathbf{f}_j^*))(\phi_*(\mathbf{f}_i^*)-\phi_*(\mathbf{f}_j^*))^T,
\label{equ:scatter_within}
\end{equation}
where $*\in\{v,d\}$, $v$ denotes visual feature and $d$ denotes depth feature, and
\begin{equation}
\mathbf{A}_{i,j}^b=
\begin{cases}
\frac {1}{N_s} - \frac{1}{N_c} \;\; \rm{if} \; \mathnormal{y_i=y_j=c}, \\
\frac{1}{N_s} \;\; \rm{if} \; \mathnormal{y_i \not= y_j},
\end{cases}
\label{equ:affinity_between}
\end{equation}
\begin{equation}
\mathbf{A}_{i,j}^w=
\begin{cases}
\frac{1}{N_c} \;\; \rm{if} \; \mathnormal{y_i=y_j=c}, \\
0 \;\; \rm{if} \; \mathnormal{y_i \not= y_j},
\end{cases}
\label{equ:affinity_within}
\end{equation}
and $N_c$ is the number of samples of class $c$.


Then we combine the minimization of $\Omega_{vd}$ with the discriminant feature learning that maximizes between-class \modifyD{variance} while minimizes within-class \modifyD{variance} as follows:
\begin{equation}
\max_{\mathbf{W}_v,\mathbf{W}_d} \frac{\gamma_0 {\rm tr}(\mathbf{W}_v^T \mathbf{S}_{bv} \mathbf{W}_v) + \gamma_1 {\rm tr}(\mathbf{W}_d^T \mathbf{S}_{bd} \mathbf{W}_d)}
{\beta^\prime \Omega_{vd} + \gamma_0^\prime {\rm tr}(\mathbf{W}_v^T \mathbf{S}_{wv} \mathbf{W}_v) + \gamma_1^\prime {\rm tr}(\mathbf{W}_d^T \mathbf{S}_{wd} \mathbf{W}_d)}
,
\label{equ:obj_fun}
\end{equation}
where $\mathbf{S}_{bv}$($\mathbf{S}_{bd}$) and $\mathbf{S}_{wv}$($\mathbf{S}_{wd}$) are between-class scatter matrix and within-class scatter matrix of visual features (depth features) \modifyC{respectively, and $\beta^\prime$, $\gamma_0$, $\gamma_0^\prime$, $\gamma_1$ and $\gamma_1^\prime$ are non-negative trade-off parameters.} \modifyB{We call the above transfer model the \emph{kernelized implicit feature transfer scheme}. It is unsupervised without using information in target domain.}

\subsection{Optimization}

We show that the model developed in the last section can be converted to a generalized eigen-decomposition problem.
As suggested by the representer theorem \cite{Bernhard2001representer}, the projection matrices can be represented by the combination of training samples, i.e., $\mathbf{W}_v=\mathbf{\Phi}_v \mathbf{A}_v$, $\mathbf{W}_d=\mathbf{\Phi}_d \mathbf{A}_d$, where $\mathbf{\Phi}_v=[\phi_v(\mathbf{f}_{1}^v),\phi_v(\mathbf{f}_{2}^v),...,\phi_v(\mathbf{f}_{N_s}^v)]\in \mathbb{R}^{m_v \times N_s}$ and $\mathbf{\Phi}_d=[\phi_d(\mathbf{f}_1^d),\phi_d(\mathbf{f}_2^d),...,\phi_d(\mathbf{f}_{N_s}^d)]\in \mathbb{R}^{m_d \times N_s}$ are visual feature matrix and depth feature matrix of training samples respectively and $\mathbf{A}_v \in \mathbb{R}^{N_s \times m}$ and $\mathbf{A}_d \in \mathbb{R}^{N_s \times m}$ are the combination coefficient matrices. For visual feature $\mathbf{f}^v$ and depth feature $\mathbf{f}^d$, we define
\begin{equation}\label{f_hat1}
\widetilde{\mathbf{f}^v}=[{\rm K}_v(\mathbf{f}_{1}^v,\mathbf{f}^v), {\rm K}_v(\mathbf{f}_{2}^v,\mathbf{f}^v),...,{\rm K}_v(\mathbf{f}_{N_s}^v,\mathbf{f}^v)]^T,
\end{equation}
\begin{equation}\label{f_hat2}
\widetilde{\mathbf{f}^d}=[{\rm K}_d(\mathbf{f}_{1}^d,\mathbf{f}^d),{\rm K}_d(\mathbf{f}_{2}^d,\mathbf{f}^d),...,{\rm K}_d(\mathbf{f}_{N_s}^d,\mathbf{f}^d)]^T,
\end{equation}
where ${\rm K}_v(\cdot,\cdot)$ and ${\rm K}_d(\cdot,\cdot)$ are kernel functions for visual feature and depth feature, respectively. The projection of a visual feature $\mathbf{f}^v$ is expressed as $\mathbf{W}_v^T \phi_v(\mathbf{f}^v)=\mathbf{A}_v^T \mathbf{\Phi}_v^T \phi_v(\mathbf{f}^v)=\mathbf{A}_v^T \widetilde{\mathbf{f}^v}$. In the same way, $\mathbf{W}_d^T \phi_d(\mathbf{f}^d)=\mathbf{A}_d^T \widetilde{\mathbf{f}^d}$. To jointly solve $\mathbf{A}_v$ and $\mathbf{A}_d$, we define $\mathbf{A}=[\mathbf{A}_v^T,\mathbf{A}_d^T]^T$, zero-padding transformation matrix $\mathbf{Z}_v=[\mathbf{I}_{N_s},\mathbf{O}_{N_s \times N_s}]$ for $\widetilde{\mathbf{f}^v}$ and $\mathbf{Z}_d=[\mathbf{O}_{N_s \times N_s},\mathbf{I}_{N_s}]$ for $\widetilde{\mathbf{f}^d}$. Now the objective function (\ref{equ:obj_fun}) can be reformulated as:
\begin{equation}
\max_{\mathbf{A}} \frac{
\gamma_0 {\rm tr}(\mathbf{A}^T \mathbf{Z}_v^T \widetilde{\mathbf{S}}_{bv} \mathbf{Z}_v \mathbf{A}) + \gamma_1 {\rm tr}(\mathbf{A}^T \mathbf{Z}_d^T \widetilde{\mathbf{S}}_{bd} \mathbf{Z}_d \mathbf{A})
}
{
\beta^\prime \widetilde{\Omega}_{vd} + \gamma_0^\prime {\rm tr}(\mathbf{A}^T \mathbf{Z}_v^T \widetilde{\mathbf{S}}_{wv} \mathbf{Z}_v \mathbf{A}) + \gamma_1^\prime {\rm tr}(\mathbf{A}^T \mathbf{Z}_d^T \widetilde{\mathbf{S}}_{wd} \mathbf{Z}_d \mathbf{A})
},
\label{equ:obj_fun2}
\end{equation}
where $\widetilde{\mathbf{S}}_{bv}$, $\widetilde{\mathbf{S}}_{wv}$, $\widetilde{\mathbf{S}}_{bd}$, $\widetilde{\mathbf{S}}_{wd}$ are scatter matrices defined by
$\widetilde{\mathbf{S}}_{b*}=\sum_{i,j=1}^{N_s} \mathbf{A}_{i,j}^b (\widetilde{\mathbf{f}_i^*}-\widetilde{\mathbf{f}_j^*})(\widetilde{\mathbf{f}_i^*}-\widetilde{\mathbf{f}_j^*})^T,
$ and $\widetilde{\mathbf{S}}_{w*}=\sum_{i,j=1}^{N_s} \mathbf{A}_{i,j}^w (\widetilde{\mathbf{f}_i^*}-\widetilde{\mathbf{f}_j^*})(\widetilde{\mathbf{f}_i^*}-\widetilde{\mathbf{f}_j^*})^T$,
 $*\in\{v,d\}$.
\begin{equation}
\begin{aligned}
\widetilde{\Omega}_{vd}&=\frac{1}{C} \sum_{c=1}^{C} \norm*{\frac{1}{N_c}\sum_{y_i=c}\mathbf{A}^T \mathbf{Z}_v^T \widetilde{\mathbf{f}}_{i}^v - \frac{1}{N_c}\sum_{y_i=c}\mathbf{A}^T \mathbf{Z}_d^T \widetilde{\mathbf{f}}_i^d}^2_2\\
&={\rm tr}(\mathbf{A}^T \mathbf{B}_{vd} \mathbf{A}),
\end{aligned}
\label{euq:MMD_vd_new}
\end{equation}
where $\mathbf{B}_{vd}=\frac{1}{C} \sum_{c=1}^{C} \mathbf{U}_c \mathbf{U}_c^T$, $\mathbf{U}_c=\frac{1}{N_c}\sum_{y_i=c}(\mathbf{Z}_v^T \widetilde{\mathbf{f}}_{i}^v - \mathbf{Z}_d^T \widetilde{\mathbf{f}}_i^d)$.

Let $\mathbf{B}_{bv}=\mathbf{Z}_v^T \widetilde{\mathbf{S}}_{bv} \mathbf{Z}_v$, $\mathbf{B}_{wv}=\mathbf{Z}_v^T \widetilde{\mathbf{S}}_{wv} \mathbf{Z}_v$, $\mathbf{B}_{bd}=\mathbf{Z}_d^T \widetilde{\mathbf{S}}_{bd} \mathbf{Z}_d$, $\mathbf{B}_{wd}=\mathbf{Z}_d^T \widetilde{\mathbf{S}}_{wd} \mathbf{Z}_d$ denote the zero-padding scatter matrices.
Finally, the objective function is formulated by:
\begin{equation}
\begin{aligned}
\max_{\mathbf{A}} \;\;
{\rm tr}(\mathbf{A}^T \mathbf{B}_{1} \mathbf{A}) \\
s.t. \;\; \mathbf{A}^T \mathbf{B}_{2} \mathbf{A}=\mathbf{I},
\end{aligned}
\label{equ:obj_fun3}
\end{equation}
where $\mathbf{B}_1= \gamma_0 \mathbf{B}_{bv} + \gamma_1 \mathbf{B}_{bd}$, $\mathbf{B}_2= \beta^\prime \mathbf{B}_{vd} + \gamma_0^\prime \mathbf{B}_{wv} + \gamma_1^\prime \mathbf{B}_{wd}$.
Hence, a generalized eigen-decomposition problem can be derived below:
\begin{equation}\label{equ:solution}
\mathbf{B}_1 \mathbf{A}=\lambda \mathbf{B}_2 \mathbf{A}.
\end{equation}
Solving the above is to compute the eigen-decomposition $\mathbf{B}_2^{-1} \mathbf{B}_1=\mathbf{A} \mathbf{\Lambda} \mathbf{A}^T$, in which $\mathbf{\Lambda}$ is a diagonal matrix with sorted eigenvalues in descending order lying on the diagonal and $\mathbf{A}\in \mathbb{R}^{2N_s\times 2N_s}$ contains the eigenvectors. Since $\mathbf{A}=[\mathbf{A}_v^T,\mathbf{A}_d^T]^T$, we can obtain $\mathbf{A}_v$ by extracting the first $N_s$ rows of $\mathbf{A}$. To specify the dimension of the common latent subspace, we use the first $m$ columns of $\mathbf{A}_v$ to form the projection matrix $\mathbf{A}_v^\prime \in \mathbb{R}^{N_s\times m}$ so that we can project visual feature to the $m$-dimensional common latent subspace.

\subsection{Depth Feature Estimation on Target Domain}
After learning the projection to the discriminative common latent subspace, we can implicitly estimate the depth feature of an RGB image in target domain by mapping visual feature to high-dimensional nonlinear space by $\phi_v(\cdot)$ and projecting it to the learned common latent subspace by $\mathbf{W}_v$. Given two new samples $p_1$ and ${p}_2$ in target domain, let $\mathbf{f}_{p1}^v$, $\mathbf{f}_{p2}^v$ denote the visual features of RGB images. The estimated depth features in the learned discriminative common latent subspace are computed by $\widehat{\mathbf{f}}_{p1}^d=\mathbf{A}_v^{\prime T} \widetilde{\mathbf{f}}_{p1}^v$ and $\widehat{\mathbf{f}}_{p2}^d=\mathbf{A}_v^{\prime T} \widetilde{\mathbf{f}}_{p2}^v$. Then the distance of depth features between $p_1$ and ${p}_2$ is computed by
\begin{equation}\label{equ:dist_D}
dist_{D}(p_1,p_2)=||\widehat{\mathbf{f}}_{p1}^d-\widehat{\mathbf{f}}_{p2}^d||_2.
\end{equation}

\section{Experiments}

Our depth-based person re-identification framework was evaluated on three RGB-D person re-identification datasets PAVIS \cite{Barbosa:reid12}, BIWI RGBD-ID \cite{Munaro:reid13} and IAS-Lab RGBD-ID \cite{Munaro:reid14}, which were captured by Kinect. In Section \ref{subsec:transfer_evaluation}, the kernelized implicit feature transfer scheme was evaluated on 3DPeS \cite{baltieri20113dpes} and CAVIAR4REID \cite{Dong2011Custom}. The experiment results were presented in Cumulative Matching Characteristic (CMC) curve \cite{moon2001computational} and rank-$k$ accuracy. Rank-$k$ accuracy is the cumulative recognition rate of correct matches at rank $k$. The CMC curve represents the cumulative recognition rates at all ranks. The evaluation was repeated 10 times and average results were reported.

\vspace{0.1cm}
\noindent \textbf{Compared Methods}. \modifyD{By following the general re-id setting,} we tested Eigen-depth feature (ED), our depth voxel covariance descriptor (DVCov), skeleton-based feature (SKL) and the combinations of depth shape descriptors and skeleton-based feature (ED+SKL and DVCov+SKL). We conducted comparisons with RGB-based appearance features including LOMO feature \cite{liao2015person}, ELF18 feature \cite{chen2015asymmetric}, color histograms (RGB, HS and YCbCr space) \cite{gray2008viewpoint}, HOG \cite{Oreifej10} and LBP \cite{lbp}, rotation invariant depth shape descriptors including RIFT2M \cite{skelly2007improved} and Fehr's covariance descriptor \cite{fehr2012compact}, and skeleton-based feature designed for depth-based re-id \cite{Munaro:reid13}. All RGB-based appearance features were extracted from images which were resized to $128\times48$. RIFT2M and Fehr's descriptor were densely extracted using the same voxels as Eigen-depth feature. \modifyB{We used LDA to learn the distance metric for all features, except that the skeleton-based feature was matched by Euclidean distance and our depth voxel covariance descriptor was matched by geodesic distance using Equation (\ref{equ:13}).}

\subsection{Evaluation on PAVIS}

We used two groups of dataset images in PAVIS dataset~\cite{Barbosa:reid12} for evaluation here. These two groups are denoted by ``Walking1'' and ``Walking2''.
Images of ``Walking1'' and ``Walking2'' were obtained by recording the same 79 people with a frontal view, walking slowly in an indoor scenario. Among the 79 people, 60 people in ``Walking2'' dressed different clothes from ``Walking1''.

\begin{figure}
\center
\includegraphics[width=0.9\linewidth]{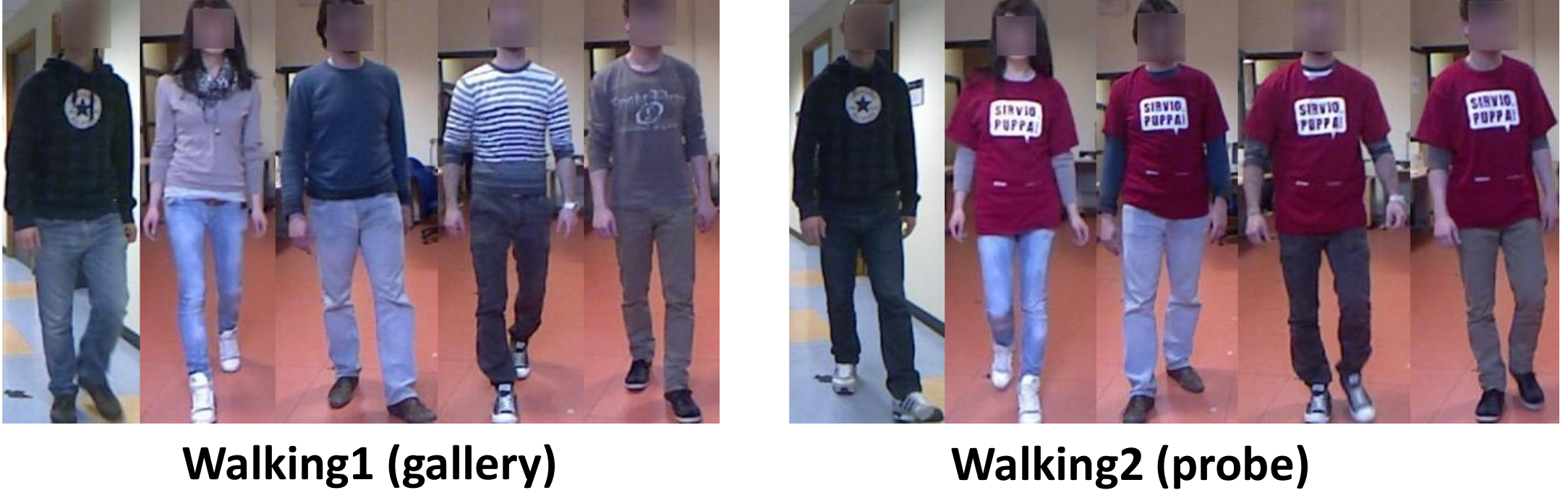}
\caption{Examples of images in ``Walking1'' and ``Walking2'' in PAVIS. Most persons in ``Walking2'' dressed different clothes from ``Walking1''.}
\label{fig:RGBD-IDimages}
\end{figure}

The characteristic of this experiment is that some people changed their clothes (by wearing one more red shirt) from ``Walking1'' to ``Walking2'' as shown in Figure \ref{fig:RGBD-IDimages}.
However, one could still explore some appearance cues (e.g., trousers and body shape) for  matching persons across these two sets. Since the images of frontal bodies were captured from nearly the same view in these two sets, there was little rotation variation of point clouds. In this case, we can apply DVCov+SKL in our framework.

\begin{figure}
\center
   \includegraphics[height=0.5\linewidth]{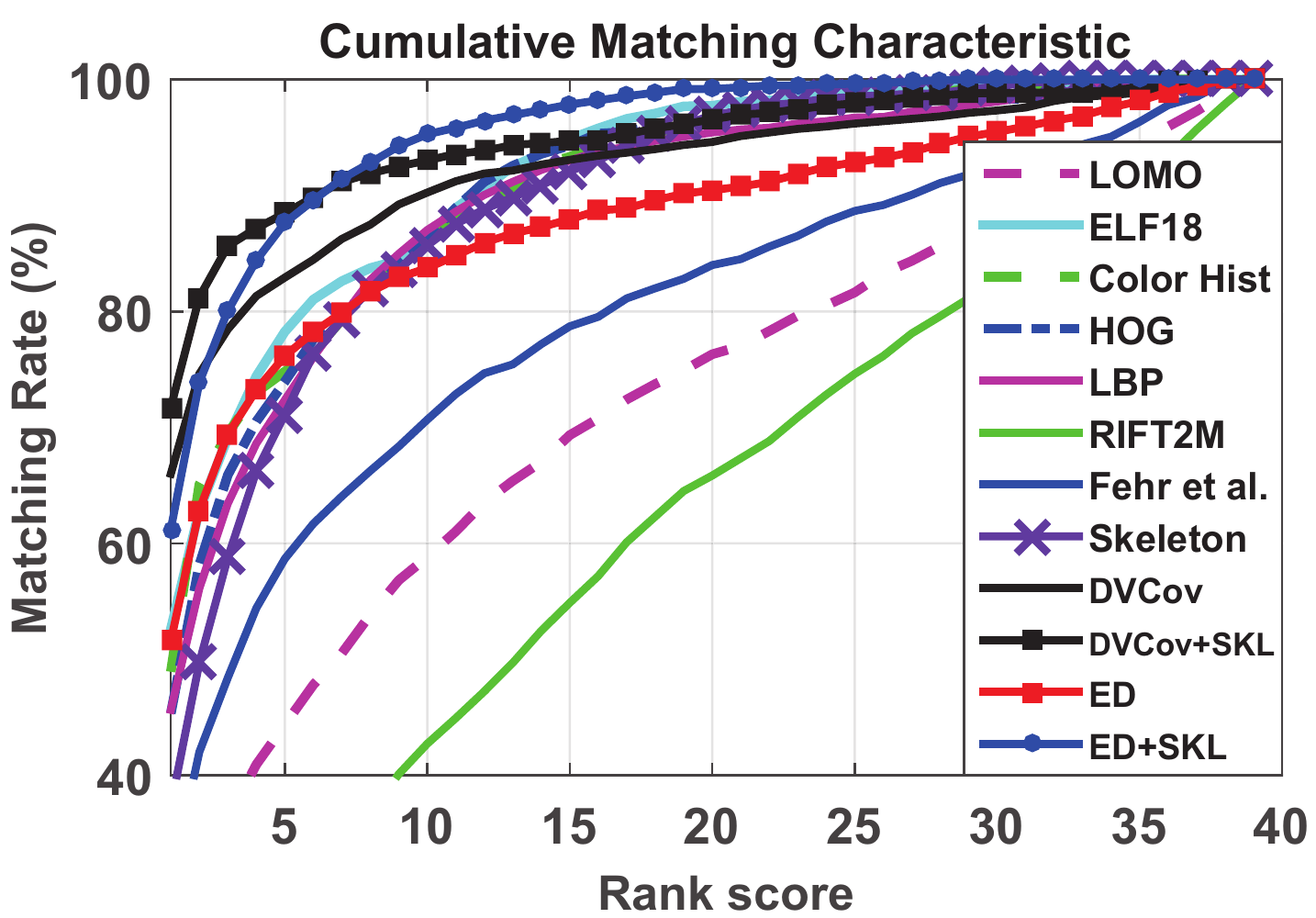}
\caption{Performance on PAVIS (multi-shot). Our approach: ED (Eigen-depth feature), ED+SKL, DVCov (our depth voxel covariance descriptor), DVCov+SKL.}
\label{fig:cmcRGBD-ID}
\end{figure}

\begin{figure*}
\center
\subfigure[PAVIS]{
   \includegraphics[width=0.3\linewidth]{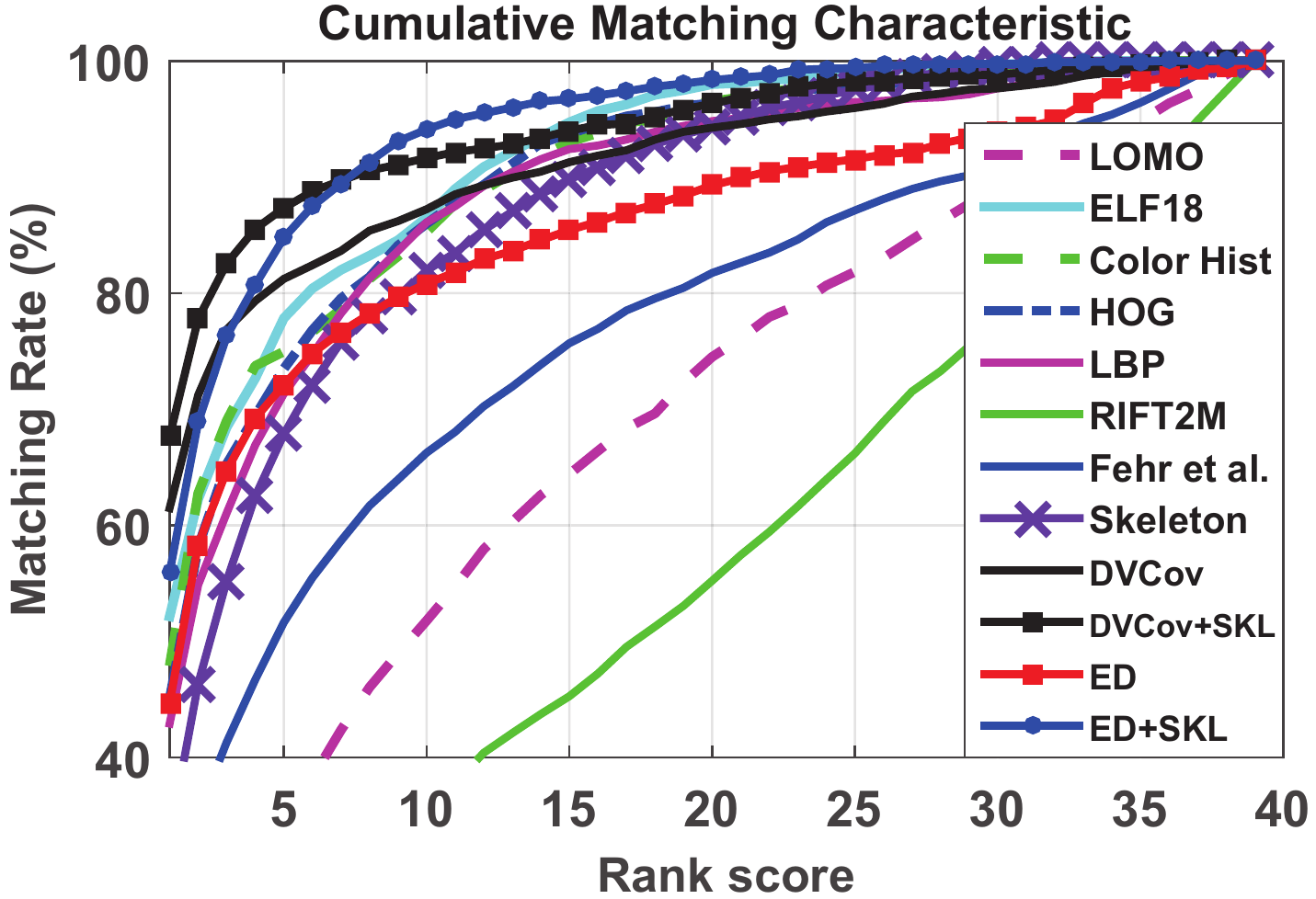}
}
\subfigure[BIWI RGBD-ID ``Still'']{
   \includegraphics[width=0.3\linewidth]{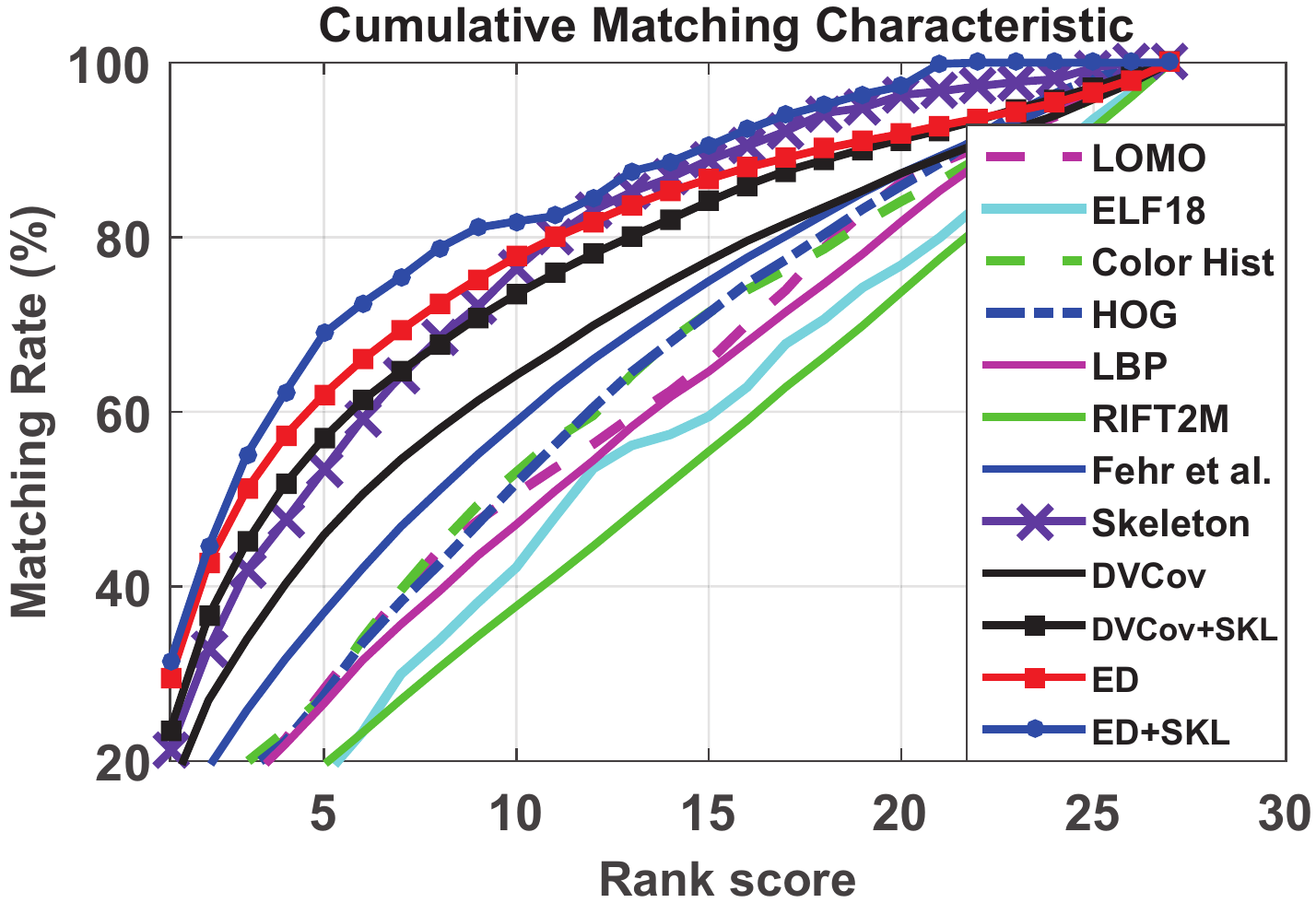}
}
\subfigure[BIWI RGBD-ID ``Walking'']{
   \includegraphics[width=0.3\linewidth]{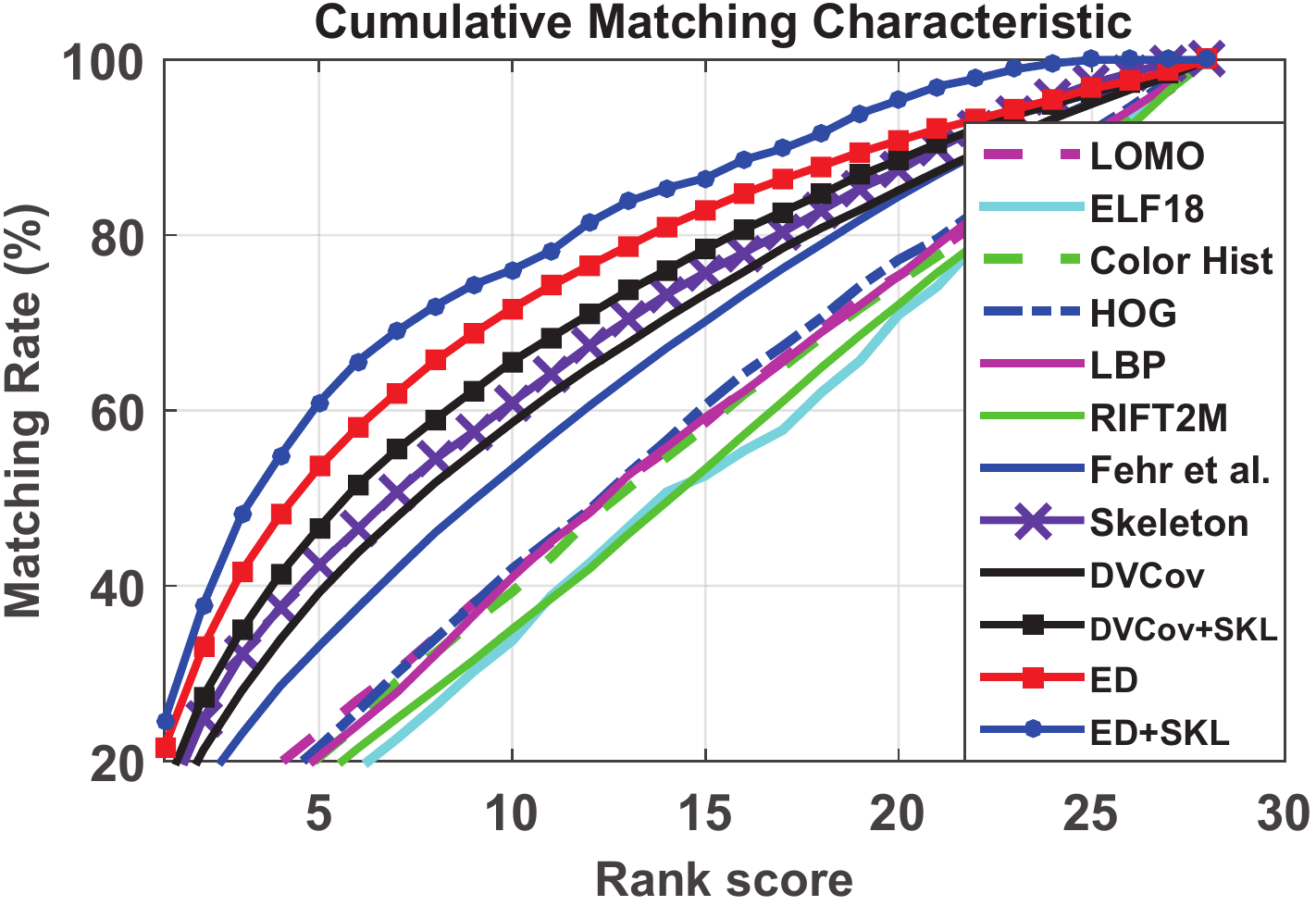}
}
\caption{Performance on PAVIS and BIWI RGBD-ID (single-shot). Our approach: ED (Eigen-depth feature), ED+SKL, DVCov (our depth voxel covariance descriptor), DVCov+SKL.}
\label{fig:cmcPAVISBIWI}
\end{figure*}

We used images in ``Walking1'' to form the gallery set and the images in ``Walking2'' to form the probe.
By following the usual train-test policy for person re-identification, we randomly sampled half of the group ``Walking1'', i.e., images of 40 persons for training, and the remaining 39 persons were used for testing. Images of these 39 testing persons in ``Walking1'' were randomly selected as gallery and all images of these 39 persons in ``Walking2'' were used as probe. In single-shot experiments, one image of each person was randomly selected as gallery. In multi-shot experiments, five images of a person were selected as gallery, and in such a case the distance between each probe image and each gallery class was the minimum distance between each probe image and each gallery image of that class.
The performance of the tested methods was reported in Figure \ref{fig:cmcRGBD-ID}, Figure \ref{fig:cmcPAVISBIWI} (a) and Table \ref{tbl:RGBD-ID}.

\begin{table}
\scriptsize
\center
\caption{PAVIS dataset: Rank-1 and Rank-5 accuracies (\%), \modifyD{including results of our proposed methods and comparisons with RGB-based appearance features and depth-based features.}}
\begin{tabular}{l|c|c|c|c}
\hline
Setting    & \multicolumn{2}{|c|}{Single-shot} & \multicolumn{2}{|c}{Multi-shot} \\
\hline
    Rank & \;\;\;\;\; 1 \;\;\;\;\; & \;\;\;\;\; 5 \;\;\;\;\; & \;\;\;\;\; 1 \;\;\;\;\; & \;\;\;\;\; 5 \;\;\;\;\; \\
\hline
\hline
\multicolumn{5}{l}{RGB-based appearance features} \\
\hline
LOMO\cite{liao2015person}       &12.05  & 35.03  & 19.74  & 44.36 \\

ELF18\cite{chen2015asymmetric}       &52.15  & 77.85  & 52.62  & 78.26  \\

Color Hist\cite{gray2008viewpoint} &    47.90 &    74.97 &    48.92 &    74.82 \\

       HOG\cite{Oreifej10} &    45.03 &    73.49 &    45.33 &    73.95 \\

       LBP\cite{lbp} &    42.92 &    71.33 &    45.64 &    72.36 \\
\hline
\hline
\multicolumn{5}{l}{Depth-based features} \\
\hline
    RIFT2M\cite{skelly2007improved} &     7.13 &    22.77 &     8.77 &    27.69 \\

    Fehr's\cite{fehr2012compact} &    24.26 &    51.64 &    30.56 &    58.67 \\

  Skeleton\cite{Munaro:reid13} &    33.13 &    67.85 &    37.33 &    71.13 \\
\hline
\hline
\multicolumn{5}{l}{Proposed} \\
\hline
DVCov (depth voxel covariance)        &      61.49 &     81.23 &     66.00 &     82.92 \\

 DVCov+SKL & {\bf 67.64} & {\bf 87.33} & {\bf 71.74} & {\bf 88.46} \\

ED (Eigen-depth feature) &      44.67  &      72.10  &      51.59  &     76.15 \\

    ED+SKL &     55.95 &     84.77 &    61.23 &    87.64 \\
\hline
\end{tabular}
\label{tbl:RGBD-ID}
\end{table}

The results suggest that both Eigen-depth feature (ED) and our depth voxel covariance descriptor (DVCov) are more effective than RIFT2M and Fehr's descriptor for describing body shape. \modifyB{Since view angles of persons are nearly the same in ``Walking1'' and ``Walking2'', our depth voxel covariance descriptor is more effective than Eigen-depth feature, because it contains richer information about textures than using only eigenvalues. However, Eigen-depth feature is still more effective than other methods except for our depth voxel covariance descriptor.} Using skeleton-based feature alone cannot achieve high performance, but it is complementary information for our depth voxel covariance descriptor and Eigen-depth feature. The combination of our depth voxel covariance descriptor and skeleton-based feature achieves encouraging performance, where the rank-1 accuracy is 67.64\% for single-shot recognition and 71.74\% for multi-shot recognition. It is clear that the fusion outperformed RGB appearance-based methods and other tested depth-based methods. \modifyD{We note that not all RGB-based appearance features performed badly as we expected, because among the 79 people, 19 people did not change clothes and the other 60 people's trousers did not change as well from the gallery set to the probe.} \modifyD{In conclusion, this test showed the effectiveness of our depth voxel covariance descriptor for shape description.}

\subsection{Evaluation on BIWI RGBD-ID}

\begin{figure}
\center
\includegraphics[width=1.0\linewidth]{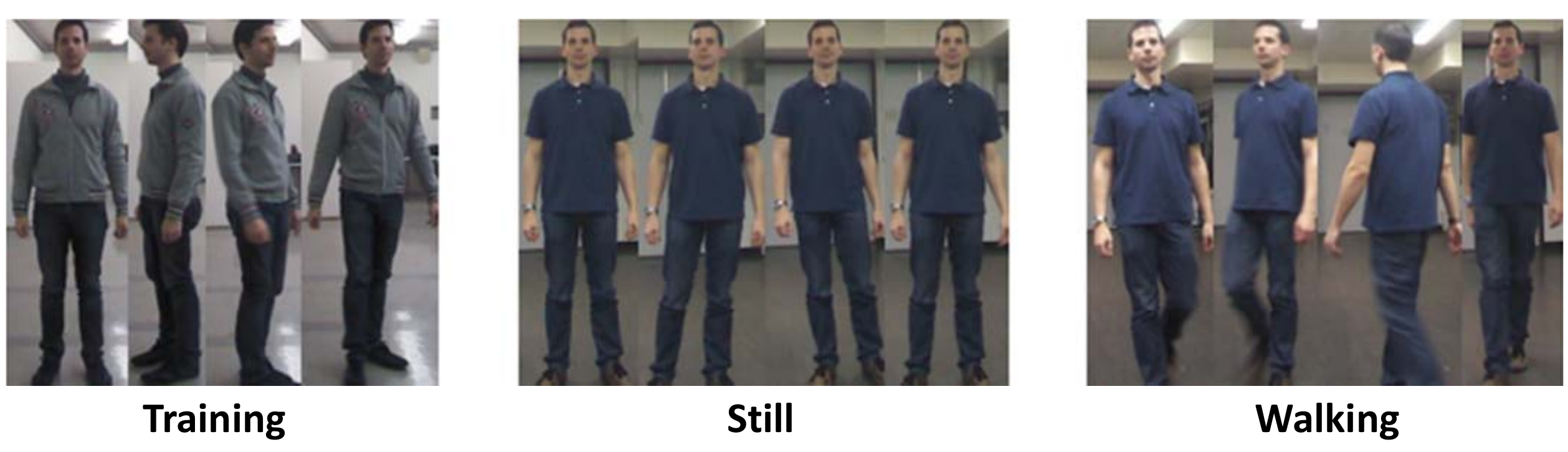}
\caption{Examples of images in BIWI RGBD-ID. \modifyD{In ``Still'', the persons were captured from frontal view, while in ``Training'' and ``Walking'' the persons were captured from multiple views.}}
\label{fig:BIWIimages}
\end{figure}

The BIWI RGBD-ID dataset~\cite{Munaro:reid13} contains three groups of sequences ``Training'', ``Still'' and ``Walking'' captured from 50 different people. For a sequence of each person, there are about 300 frames of depth images and skeletons. Before feature extraction, we converted depth images to point clouds. In ``Training'', people performed motions, such as walking and rotating. Only 28 people presented in ``Training'' were recorded in ``Still'' and ``Walking'', which were collected in a different day and in a different scene, so that most persons were dressed differently. In ``Still'', people slightly moved, while in ``Walking'', every person walked in different view angles. Examples of images in ``Training'', ``Still'' and ``Walking'' are shown in Figure \ref{fig:BIWIimages}. Since pedestrians' viewpoint variation was large here, it was more suitable to use ED+SKL in our framework.

For BIWI RGBD-ID dataset, images of the 22 people who only appeared in ``Training'' were used for training, and images of the remaining 28 people were used for testing. In the testing set, we used images in ``Training'' as gallery and images in ``Still'' and ``Walking'' as probe, so the same person wore different clothes in gallery and probe.
\modifyD{We selected the samples for evaluation by face detection as advised in \cite{Munaro:reid13}.} Since the persons were captured from different view angles, this dataset is suitable to evaluate the effect of the local rotation invariance property of the proposed Eigen-depth feature.
The average results of CMC curve and rank-$k$ accuracy over 10 trials were reported in Figure \ref{fig:cmcPAVISBIWI} (b), (c) and Table \ref{tbl:Still&Walking}.

\begin{table}
\scriptsize
\center
\caption{BIWI RGBD-ID dataset ``Still'' and ``Walking'': Rank-1 and Rank-5 accuracies (\%), \modifyD{including results of our proposed methods and comparisons with RGB-based appearance features and depth-based features}.}
\begin{tabular}{l|c|c|c|c|c|c|c|c}
\hline
Probe       & \multicolumn{4}{c|}{Still}                            & \multicolumn{4}{c}{Walking} \\
\hline
Setting     & \multicolumn{2}{c|}{Single-shot} & \multicolumn{2}{c|}{Multi-shot} & \multicolumn{2}{c|}{Single-shot} & \multicolumn{2}{c}{Multi-shot} \\
\hline
Rank      & \;\;\;\;\;1\;\;\;\;\; & \;\;\;\;\;5\;\;\;\;\; & \;\;\;\;\;1\;\;\;\;\; & \;\;\;\;\;5\;\;\;\;\; & \;\;\;\;\;1\;\;\;\;\; & \;\;\;\;\;5\;\;\;\;\; & \;\;\;\;\;1\;\;\;\;\; & \;\;\;\;\;5\;\;\;\;\; \\
\hline
\multicolumn{9}{l}{RGB-based appearance features} \\
\hline
LOMO\cite{liao2015person}  &9.07  & 28.21  & 18.17  & 35.47  & 8.74  & 23.33  & 10.31  & 25.39\\

ELF18\cite{chen2015asymmetric}       &2.79  & 18.18  & 4.11  & 19.13  & 1.32  & 16.03  & 1.50  & 16.77\\

Color Hist\cite{gray2008viewpoint}  & 7.02      & 25.47     & 10.61     & 31.92     & 5.43      & 19.56     & 5.86      & 21.70 \\

HOG\cite{Oreifej10}         & 8.42      & 25.69     & 12.35     & 30.39     & 6.38      & 21.00     & 6.94      & 23.29 \\

LBP\cite{lbp}         & 7.37      & 26.04     & 10.87     & 33.57     & 4.87      & 20.04     & 5.34      & 23.31 \\
\hline
\hline
\multicolumn{9}{l}{Depth-based features} \\
\hline
RIFT2M\cite{skelly2007improved}      & 4.04      & 19.52     & 4.34      & 20.78     & 3.25      & 17.46     & 3.75      & 18.31 \\

Fehr's\cite{fehr2012compact}    & 12.08     & 38.17     & 14.06     & 43.78     & 9.33      & 32.39     & 12.09     & 39.60 \\

Skeleton\cite{Munaro:reid13}    & 21.34     & 53.32     & 26.55     & 62.73     & 14.52     & 42.36     & 16.94     & 47.18 \\
\hline
\hline
\multicolumn{9}{l}{Proposed} \\
\hline
DVCov         & 16.32     & 45.93     & 23.07     & 58.89     & 12.58     & 39.22     & 17.24     & 45.93 \\

DVCov+SKL     & 23.49     & 57.06     & 34.37     & 72.77     & 16.59     & 46.67     & 21.40     & 54.12 \\

ED          & 28.98     & 61.85     & 36.22     & \textbf{73.11} & 20.90     & 51.98     & 28.71     & 63.85 \\

ED+SKL      & \textbf{30.52} & \textbf{67.86} & \textbf{39.38} & 72.13     & \textbf{24.47} & \textbf{60.63} & \textbf{29.96} & \textbf{65.18} \\
\hline
\end{tabular}%
\label{tbl:Still&Walking}

\end{table}

As shown in Figure \ref{fig:cmcPAVISBIWI} (b), (c), RGB-based appearance features completely failed, because most people changed clothes so that color feature was not reliable. Our depth-based methods outperformed all RGB appearance-based methods. On BIWI RGBD-ID, people appeared in different view angles, so the problem became more challenging than the one on PAVIS. On ``Walking'', the problem was even more difficult since more frames were captured in multiple viewpoints. In these situations, rotation invariant depth shape descriptor is more suitable, so that our Eigen-depth feature outperformed our depth voxel covariance descriptor. Compared with other rotation invariant depth shape descriptors, Eigen-depth feature outperformed RIFT2M and Fehr's descriptor.
The combination of Eigen-depth feature and skeleton-based feature (ED+SKL) can achieve better performance than using them separately, \modifyD{which is the best on BIWI RGBD-ID}. \modifyD{The results showed the local rotation invariance and the effectiveness of body shape description of Eigen-depth feature.}

\subsection{Evaluation on IAS-Lab RGBD-ID}

\begin{figure}
\center
\includegraphics[width=1.0\linewidth]{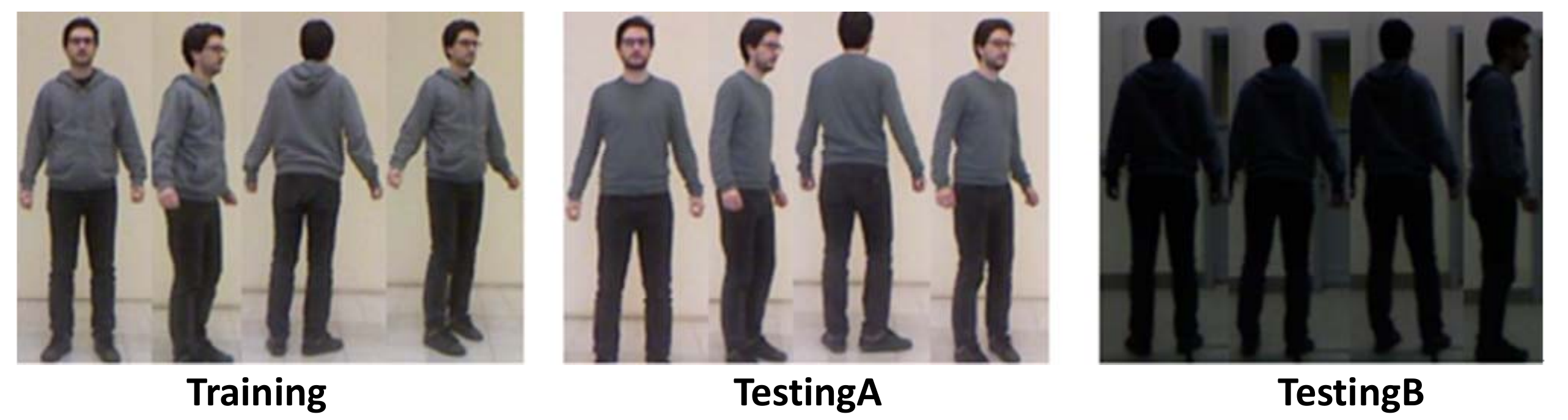}
\caption{Examples of images in IAS-Lab RGBD-ID. \modifyD{All samples were captured from multiple views. Compared to ``Training'', samples in ``TestingA'' changed clothes and some samples in ``TestingB'' were captured in dark environment.}}
\label{fig:IASimages}
\end{figure}

There are 11 different people in IAS-Lab RGBD-ID dataset \cite{Munaro:reid14}. In this dataset, three groups of sequences ``Training'', ``TestingA'' and ``TestingB'' were recorded, and each person rotated on himself and walked during the recording. There are about 500 frames of depth images and skeletons for each person. The sequences in ``Training'' and ``TestingA'' were acquired when the same person was wearing different clothes. The sequences in ``TestingB'' were collected in a different room, where each person dressed the same as in ``Training''. Some sequences in ``TestingB'' were recorded in dark environment. Examples of images in ``Training'', ``TestingA'' and ``TestingB'' are shown in Figure \ref{fig:IASimages}.

On this dataset, the evaluation also followed the settings on PAVIS. Half of ``Training'' sequences were randomly selected to form the training set and the rest were selected to form the gallery in the test. \modifyC{The samples in ``TestingA'' and ``TestingB'' corresponding to the gallery persons were selected to form the probe set.} By following the settings in \cite{Munaro:reid14}, all images were used in this experiment. On this dataset, mismatch would be observed when performing the matching between a person image of rear view and his/her image of frontal view, so that it challenges body shape descriptors. The average rank-1 and rank-3 accuracies over 10 trials of evaluation were reported in Table \ref{tbl:AB}.

\begin{table}
\scriptsize
\center
\caption{IAS-Lab RGBD-ID dataset ``TestingA'' and ``TestingB'': Rank-1 and Rank-3 accuracy (\%), \modifyD{including results of our proposed methods and comparisons with RGB-based appearance features and depth-based features}.}
\begin{tabular}{l|c|c|c|c|c|c|c|c}
\hline
Probe      & \multicolumn{4}{|c|}{TestingA} & \multicolumn{4}{|c}{TestingB} \\
\hline
    Setting &   \multicolumn{2}{|c|}{Single-shot} & \multicolumn{2}{|c|}{Multi-shot} &   \multicolumn{2}{|c|}{Single-shot} & \multicolumn{2}{|c}{Multi-shot} \\
\hline
    Rank    & ~~~~1~~~~ & ~~~~3~~~~ & ~~~~1~~~~ & ~~~~3~~~~ & ~~~~1~~~~ & ~~~~3~~~~ & ~~~~1~~~~ & ~~~~3~~~~\\
\hline
\multicolumn{5}{l}{RGB-based appearance features} \\
\hline
LOMO\cite{liao2015person} &26.37&65.82&25.79&66.28&30.97&75.00&30.06&79.90\\

ELF18\cite{chen2015asymmetric} &22.35& 60.96 & 21.81& 67.77 & 24.03& 67.36 & 23.01&67.81\\

Color Hist\cite{gray2008viewpoint} &27.69&63.71&    24.42&66.48&    18.45&63.33&    23.89&60.93\\

       HOG\cite{Oreifej10} &    31.00&66.48&    38.89&72.67&    47.21&81.16&    49.62&86.79\\

       LBP\cite{lbp} &28.71&67.97&32.81&68.22&    51.38&84.28&52.88&89.81\\

\hline
\hline
\multicolumn{5}{l}{Depth-based features} \\
\hline
    RIFT2M\cite{skelly2007improved} &    19.69&60.76&    20.94&60.87&    19.88&59.78&    19.88&60.02\\

    Fehr's\cite{fehr2012compact} &    23.78&67.34&    24.05&64.95&    20.58&63.21&    20.46&62.65\\

  Skeleton\cite{Munaro:reid13} &    41.36&85.29&    49.83&{\bf 91.49}&     54.18&87.07&    60.25&{\bf 93.58}\\
\hline
\hline
\multicolumn{5}{l}{Proposed} \\
\hline
DVCov       &27.95&67.20&35.56&72.53&      25.38&59.67 &36.14&71.45\\

 DVCov+SKL &34.10&71.00&46.57&79.23&27.74&62.28&      45.91&80.42\\

ED          &     32.09&75.23&     31.76&75.15&      35.82&73.60&      39.20&79.86 \\

    ED+SKL & {\bf 48.75}&{\bf 90.57}& {\bf 52.30}& 90.15 &  {\bf 58.65}&{\bf 94.36}& {\bf 63.29}& 91.21 \\
\hline
\end{tabular}
\label{tbl:AB}
\end{table}

On ``TestingA'', the RGB-based appearance features nearly failed, and Eigen-depth feature and skeleton-based feature outperformed them. On ``TestingB'', HOG and LBP can still adapt to illumination change to some extent, while color histogram completely failed. Since rotation of samples took place in this dataset, the proposed Eigen-depth feature outperformed our depth voxel covariance descriptor and is more suitable for shape description in such a situation. Eigen-depth feature also outperformed the compared rotation invariant depth shape descriptors RIFT2M and Fehr's descriptor. In most cases, combining Eigen-depth feature with skeleton-based feature worked better than using them separately. Since viewpoint of pose changed from 0\degree~to 360\degree~for each person in the training and testing sets, shape description from front to back for the same person changes notably and thus would cause confusion for matching. Skeleton-based feature is more effective in the cases when there are only 5 persons in testing set, because there are fewer persons of similar somatotype. So skeleton-based feature is better than Eigen-depth feature in this case. In general, the combination of Eigen-depth feature and skeleton-based feature is the most effective. \modifyD{The test showed the effectiveness of our depth-based method when people change clothes and appear in the extreme lighting condition.}

\subsection{Comparison to Depth-based Re-id Frameworks}
\label{close-set}
\modifyD{Existing well-known methods related to depth-based person re-identification include still-image-based recurrent attention model (3D RAM) \cite{feifei2016CVPR}, skeleton-based feature (SKL), Point Cloud Matching (PCM) and the combination of PCM and SKL (PCM+SKL) \cite{Munaro:reid14}. 3D RAM, PCM and PCM+SKL are designed under a different setting from the usual one for person re-id; that is they require that the group of persons  for training is the same as the one of persons for testing, while there is no overlap on persons between training and testing in the usual re-id setting. To compare our method with the above methods, we tested our Eigen-depth (ED) feature and the combination of Eigen-depth feature and skeleton-based feature (ED+SKL) on PAVIS and IAS-Lab RGBD-ID under the same setting as the compared methods when they were reported in \cite{feifei2016CVPR,Munaro:reid14}. 
The experiment results were reported in Table \ref{tbl:last1}. As shown, our method ED and ED+SKL clearly outperformed other existing depth-based frameworks, especially on PAVIS, a much larger dataset with more persons involved. 
}

\begin{table}
\scriptsize
\center
\caption{PAVIS and IAS-Lab RGBD-ID*: \modifyD{results of comparisons with existing depth-based re-id frameworks (\%).}}
\begin{tabular}{c|c|cc|cccc}
\hline
Dataset & Probe & ~ED~    & ~ED+SKL~ & 3D RAM\cite{feifei2016CVPR} & PCM\cite{Munaro:reid14} & PCM+SKL \cite{Munaro:reid14}  & SKL\cite{Munaro:reid14}\\
\hline
PAVIS & Walking2 & 54.4  & \textbf{57.0} & 41.3  & -     & - & 28.6\\
\hline
IAS-Lab & TestingA & 44.0  & \textbf{49.9} & 48.3  & 28.6  & 25.6 & 22.5\\
RGBD-ID & TestingB & 55.5  & \textbf{66.6} & 63.7  & 43.7  & 63.3 & 55.5\\
\hline
\end{tabular}%
\label{tbl:last1}
\scriptsize{\\~*The experiments here are under a different setting from the experiments in previous sections. See Sec. \ref{close-set} for details}.
\end{table}

\subsection{Depth Feature Transfer Evaluation}
\label{subsec:transfer_evaluation}

The effectiveness of the kernelized implicit feature transfer scheme was evaluated on RGB datasets 3DPeS \cite{baltieri20113dpes} and CAVIAR4REID \cite{Dong2011Custom}. Before showing the experiment results, we first \modifyC{present} implementation details of the feature transfer scheme.

\vspace{0.1cm}
\noindent \textbf{Implementation Details}. In this work, we selected the BIWI RGBD-ID dataset \cite{Munaro:reid13} as \modifyC{the} auxiliary dataset. In ``Training'' of BIWI RGBD-ID, there were 50 persons performing actions of rotation and walking. For each of the 50 persons in ``Training'', 8 RGB images from 8 different views ranged from 0\degree~to 360\degree~were selected as auxiliary RGB images. \modifyB{As for depth information, for each person, 8 point clouds from frontal view were selected for extracting depth features corresponding to those 8 RGB images.} Some samples of auxiliary RGB-D dataset are shown in Figure \ref{fig:transfer_pipeline}.

After constructing the RGB-D auxiliary dataset, we extracted visual features and depth features to establish the connection between two modalities by the kernelized implicit feature transfer scheme. Since depth features describe body shape of pedestrians, the visual features for learning the transformation should also be able to describe body shape to some extent. We used HOG \cite{Oreifej10} and LBP \cite{lbp} for describing body silhouette and textures. All RGB images in auxiliary dataset were resized to $128\times48$ for extracting HOG and LBP features using $8\times8$ cells. We also extracted the same visual features for samples in target domain. As for the point clouds, Eigen-depth feature was extracted to describe body shape.

With the extracted visual features and depth features, we \modifyD{conducted the proposed} kernelized implicit feature transfer scheme. We chose the guassian kernel functions for visual feature and depth feature, which are ${\rm K}_v(\mathbf{x},\mathbf{y})={\rm exp}(-\gamma_v||\mathbf{x}-\mathbf{y}||^2)$ and ${\rm K}_d(\mathbf{x},\mathbf{y})={\rm exp}(-\gamma_d||\mathbf{x}-\mathbf{y}||^2)$, respectively. Let $dist_{vm}$ and $dist_{dm}$ denote the means of the distances of visual features and depth features between any two samples in the auxiliary dataset, respectively. We set the bandwidth parameters as $\gamma_v=\frac{1}{dist_{vm}^2}$ and $\gamma_d=\frac{1}{dist_{dm}^2}$. As for the parameters setting of the objective function, we empirically set the default parameters as $\beta^\prime=\frac{10}{{\rm tr}(\mathbf{B}_{vd})}$, $\gamma_1=\frac{10}{{\rm tr}(\mathbf{B}_{bd})}$, $\gamma_1^\prime=\frac{10}{{\rm tr}(\mathbf{B}_{wd})}$, $\gamma_0=\frac{1}{{\rm tr}(\mathbf{B}_{bv})}$, $\gamma_0^{\prime}=\frac{1}{{\rm tr}(\mathbf{B}_{wv})}$ which were normalized by traces. That is to say, the terms related to depth features were assigned much larger weights since we focused on learning the relation between depth feature and visual feature in order to take advantage of the discriminative information in depth features. As for the dimension of the common latent subspace, we set $m=700$.

\vspace{0.1cm}
\noindent \textbf{Score-level Feature Fusion}. We estimated depth features on RGB images in order to augment the visual features with complementary information in depth features. Let $dist_{RGB}(p_1,p_2)$ denote the distance between RGB-based appearance features of RGB images between two samples $p_1$ and $p_2$, and $dist_{D}(p_1,p_2)$ denote the distance between depth features computed according to Equation (\ref{equ:dist_D}). We fused these two types of distances with a weight $\eta$ as follow:
\begin{equation}\label{equ:dist_fusion}
dist_{fusion}({p}_1,{p}_2)=(1-\eta)dist_{RGB}({p}_1,{p}_2)+\eta dist_{D}({p}_1,{p}_2).
\end{equation}
In our experiments, each type of distance was normalized by its mean distance between any two samples in training set.

\vspace{0.5cm}
\noindent \textbf{Experiment Settings}. We evaluated how the transferred Eigen-depth feature (TED) can help to improve the performance when combined with LOMO \cite{liao2015person} and ELF18 \cite{chen2015asymmetric}, which were two recently proposed effective RGB-based appearance features in person re-identification. To compute the similarity of RGB-based appearance features, we applied three favorable distance metric learning methods LFDA \cite{pedagadi2013local}, MLAPG \cite{LiaoICCV2015Efficient} and KISSME \cite{kostinger2012large}. So we had the following different settings, ELF18(LFDA)+TED, LOMO(LFDA)+TED, ELF18(MLAPG) +TED, LOMO(MLAPG)+TED, ELF18(KISSME)+TED, LOMO(KISSME)+TED. \modifyD{For these settings,} the corresponding default distance fusion weight $\eta$ was set to 0.3, 0.2, 0.3, 0.15, 0.3, 0.2, respectively. \modifyC{It is reasonable that the distance fusion weight $\eta$ was set to different values when fusing different RGB-based distance metrics with the depth one.} Experiments were conducted on 3DPeS and CAVIAR4REID. We followed the experiment settings on PAVIS. For each person in testing set, one image was randomly selected as gallery and the remaining images were used for probing.

\modifyF{
As for baseline methods, CCA \cite{weenink2003canonical} and sparse regression \cite{lee2006efficient} were compared.
In details, we used CCA to maximize the correlation between RGB feature and depth feature on the auxiliary dataset. As for sparse regression, we made the sparse representation shared between RGB and depth feature dictionaries so as to derive a transferred depth feature. The depth features transferred by CCA and sparse regression are denoted by D-CCA and D-SPA, respectively. We combined the distance of the transferred depth feature with the distance of RGB-based appearance feature for recognition.
}
The average rank-1 to rank-5 accuracies over 10 trials were reported in Table VII.

\vspace{0.1cm}
\noindent \textbf{Results}. The transferred Eigen-depth feature (TED) can achieve rank-1 accuracy 16.0\% on 3DPeS and 27.8\% on CAVIAR4REID. For all RGB-based appearance features and distance metrics in our experiments, TED is effective for improving the top-rank matching accuracies. The augmentation of TED can boost rank-1 accuracy of ELF18 using LFDA metric by 4.4\% on both 3DPeS and CAVIAR4REID.
Although LOMO is a state-of-the-art feature for person re-identification, the transferred depth feature makes consistent improvement especially at the rank-1 matching case.
\modifyF{
Compared to the baseline methods, the proposed implicit feature transfer scheme clearly outperformed CCA (D-CCA) and sparse regression (D-SPA) when applied for the same purpose. The results indicate that it may not be effective to use CCA and sparse regression to exploit transferred depth feature.}
Overall, it is evident that the transferred Eigen-depth feature (TED) is complementary to RGB color and texture features, so that it can augment the RGB feature representation and help to get better ranking results.

\begin{table}
\scriptsize
  \centering
  \modifyF{
  \caption{3dpes and caviar4reid: Evaluation of the transferred depth feature when combined with RGB-based appearance features using different metrics (\%).}
  }
\begin{tabular}{l|c|c|c|c|c|c|c|c|c|c}
\hline
Dataset & \multicolumn{5}{c|}{3DPeS}            & \multicolumn{5}{c}{CAVIAR4REID} \\
\hline
Rank  & \multicolumn{1}{c|}{~~~1~~~} & \multicolumn{1}{c|}{~~~2~~~} & \multicolumn{1}{c|}{~~~3~~~} & \multicolumn{1}{c|}{~~~4~~~} & \multicolumn{1}{c|}{~~~5~~~} & \multicolumn{1}{c|}{~~~1~~~} & \multicolumn{1}{c|}{~~~2~~~} & \multicolumn{1}{c|}{~~~3~~~} & \multicolumn{1}{c|}{~~~4~~~} & \multicolumn{1}{c}{~~~5~~~} \\
\hline
TED   & \textbf{16.0} & \textbf{21.7} & \textbf{26.4} & \textbf{29.0} & \textbf{32.1} & \textbf{27.8} & \textbf{35.2} & \textbf{39.7} & \textbf{43.6} & \textbf{46.8} \\

 \modifyF{D-CCA} & 6.5   & 11.0  & 15.0  & 17.8  & 20.5  & 24.2  & 31.9  & 36.6  & 40.1  & 42.9  \\
\modifyF{D-SPA} & 2.7   & 4.5   & 6.4   & 7.5   & 9.0   & 5.7   & 9.2   & 11.8  & 14.7  & 17.9  \\

\hline
\hline
\multicolumn{11}{l}{LFDA metric} \\
\hline
ELF18 & \multicolumn{1}{c|}{30.3 } & \multicolumn{1}{c|}{40.5 } & \multicolumn{1}{c|}{46.4 } & \multicolumn{1}{c|}{51.5 } & \multicolumn{1}{c|}{55.3 } & \multicolumn{1}{c|}{32.6 } & \multicolumn{1}{c|}{42.9 } & \multicolumn{1}{c|}{49.6 } & \multicolumn{1}{c|}{55.2 } & \multicolumn{1}{c}{59.0 } \\
ELF18+TED & \multicolumn{1}{c|}{\textbf{34.7 }} & \multicolumn{1}{c|}{\textbf{45.2 }} & \multicolumn{1}{c|}{\textbf{51.3 }} & \multicolumn{1}{c|}{\textbf{56.5 }} & \multicolumn{1}{c|}{\textbf{60.0 }} & \multicolumn{1}{c|}{\textbf{37.0 }} & \multicolumn{1}{c|}{\textbf{45.8 }} & \multicolumn{1}{c|}{\textbf{52.0 }} & \multicolumn{1}{c|}{\textbf{56.9 }} & \multicolumn{1}{c}{\textbf{60.8 }} \\
\modifyF{ELF18+D-CCA} & 30.0  & 40.5  & 47.2  & 52.7  & 56.7  & 35.7  & 44.0  & 48.9  & 53.3  & 56.6  \\
\modifyF{ELF18+D-SPA} & 30.3  & 40.5  & 47.1  & 51.5  & 55.6  & 32.2  & 41.9  & 47.6  & 53.0  & 57.5  \\
\hline
LOMO  & \multicolumn{1}{c|}{41.4 } & \multicolumn{1}{c|}{53.4 } & \multicolumn{1}{c|}{60.4 } & \multicolumn{1}{c|}{64.3 } & \multicolumn{1}{c|}{68.0 } & \multicolumn{1}{c|}{40.2 } & \multicolumn{1}{c|}{50.1 } & \multicolumn{1}{c|}{56.7 } & \multicolumn{1}{c|}{61.8 } & \multicolumn{1}{c}{65.6 } \\
LOMO+TED & \multicolumn{1}{c|}{\textbf{43.8 }} & \multicolumn{1}{c|}{\textbf{54.9 }} & \multicolumn{1}{c|}{\textbf{61.2 }} & \multicolumn{1}{c|}{\textbf{65.7 }} & \multicolumn{1}{c|}{\textbf{68.8 }} & \multicolumn{1}{c|}{\textbf{42.2 }} & \multicolumn{1}{c|}{\textbf{51.4 }} & \multicolumn{1}{c|}{\textbf{56.9 }} & \multicolumn{1}{c|}{\textbf{62.1 }} & \multicolumn{1}{c}{\textbf{65.7 }} \\
\modifyF{LOMO+D-CCA} & 41.2  & 52.8  & 59.6  & 64.0  & 67.2  & 40.9  & 49.3  & 54.9  & 59.5  & 63.3  \\
\modifyF{LOMO+D-SPA} & 40.2  & 51.8  & 59.2  & 63.8  & 66.9  & 38.8  & 47.9  & 53.8  & 59.0  & 62.7  \\
\hline
\hline
\multicolumn{11}{l}{MLAPG metric} \\
\hline
ELF18 & \multicolumn{1}{c|}{35.5 } & \multicolumn{1}{c|}{47.1 } & \multicolumn{1}{c|}{54.2 } & \multicolumn{1}{c|}{59.1 } & \multicolumn{1}{c|}{62.8 } & \multicolumn{1}{c|}{34.5 } & \multicolumn{1}{c|}{46.4 } & \multicolumn{1}{c|}{54.0 } & \multicolumn{1}{c|}{60.0 } & \multicolumn{1}{c}{65.1 } \\
ELF18+TED & \multicolumn{1}{c|}{\textbf{38.6 }} & \multicolumn{1}{c|}{\textbf{49.7 }} & \multicolumn{1}{c|}{\textbf{56.6 }} & \multicolumn{1}{c|}{\textbf{61.8 }} & \multicolumn{1}{c|}{\textbf{65.1 }} & \multicolumn{1}{c|}{\textbf{38.5 }} & \multicolumn{1}{c|}{\textbf{49.3 }} & \multicolumn{1}{c|}{\textbf{55.7 }} & \multicolumn{1}{c|}{\textbf{60.8 }} & \multicolumn{1}{c}{\textbf{65.7 }} \\
\modifyF{ELF18+D-CCA} & 33.9  & 45.9  & 52.3  & 57.5  & 61.6  & 36.9  & 46.1  & 52.1  & 56.8  & 60.3  \\
\modifyF{ELF18+D-SPA} & 34.5  & 46.7  & 53.3  & 59.1  & 62.9  & 34.2  & 45.0  & 52.2  & 58.4  & 62.5  \\
\hline
LOMO  & \multicolumn{1}{c|}{47.1 } & \multicolumn{1}{c|}{58.5 } & \multicolumn{1}{c|}{64.5 } & \multicolumn{1}{c|}{68.5 } & \multicolumn{1}{c|}{71.7 } & \multicolumn{1}{c|}{40.6 } & \multicolumn{1}{c|}{51.8 } & \multicolumn{1}{c|}{59.4 } & \multicolumn{1}{c|}{65.2 } & \multicolumn{1}{c}{69.4 } \\
LOMO+TED & \multicolumn{1}{c|}{\textbf{48.4 }} & \multicolumn{1}{c|}{\textbf{58.7 }} & \multicolumn{1}{c|}{\textbf{64.6 }} & \multicolumn{1}{c|}{\textbf{68.8 }} & \multicolumn{1}{c|}{\textbf{72.0 }} & \multicolumn{1}{c|}{\textbf{42.8 }} & \multicolumn{1}{c|}{\textbf{52.9 }} & \multicolumn{1}{c|}{\textbf{59.8 }} & \multicolumn{1}{c|}{\textbf{65.3 }} & \multicolumn{1}{c}{\textbf{69.6 }} \\
\modifyF{LOMO+D-CCA} & 43.7  & 55.2  & 62.3  & 66.5  & 69.7  & 41.6  & 50.0  & 56.3  & 60.9  & 64.8  \\
\modifyF{LOMO+D-SPA} & 44.3  & 55.8  & 62.3  & 66.5  & 69.3  & 39.0  & 48.8  & 55.3  & 60.9  & 65.2  \\
\hline
\hline
\multicolumn{11}{l}{KISSME metric} \\
\hline
ELF18 & \multicolumn{1}{c|}{32.4 } & \multicolumn{1}{c|}{42.8 } & \multicolumn{1}{c|}{48.9 } & \multicolumn{1}{c|}{53.5 } & \multicolumn{1}{c|}{57.0 } & \multicolumn{1}{c|}{33.3 } & \multicolumn{1}{c|}{42.6 } & \multicolumn{1}{c|}{48.7 } & \multicolumn{1}{c|}{53.5 } & \multicolumn{1}{c}{57.7 } \\
ELF18+TED & \multicolumn{1}{c|}{\textbf{35.3 }} & \multicolumn{1}{c|}{\textbf{45.4 }} & \multicolumn{1}{c|}{\textbf{52.1 }} & \multicolumn{1}{c|}{\textbf{56.7 }} & \multicolumn{1}{c|}{\textbf{59.8 }} & \multicolumn{1}{c|}{\textbf{36.3 }} & \multicolumn{1}{c|}{\textbf{45.6 }} & \multicolumn{1}{c|}{\textbf{50.9 }} & \multicolumn{1}{c|}{\textbf{55.3 }} & \multicolumn{1}{c}{\textbf{59.5 }} \\
\modifyF{ELF18+D-CCA} & 32.6  & 42.6  & 49.7  & 54.4  & 57.8  & 35.9  & 43.7  & 48.4  & 52.8  & 56.0  \\
\modifyF{ELF18+D-SPA} & 32.4  & 42.6  & 48.5  & 53.5  & 57.0  & 33.1  & 42.1  & 47.8  & 52.4  & 56.7  \\
\hline
LOMO  & \multicolumn{1}{c|}{44.3 } & \multicolumn{1}{c|}{54.6 } & \multicolumn{1}{c|}{61.3 } & \multicolumn{1}{c|}{65.2 } & \multicolumn{1}{c|}{68.7 } & \multicolumn{1}{c|}{42.7 } & \multicolumn{1}{c|}{52.7 } & \multicolumn{1}{c|}{59.4 } & \multicolumn{1}{c|}{64.3 } & \multicolumn{1}{c}{69.0 } \\
LOMO+TED & \multicolumn{1}{c|}{\textbf{45.2 }} & \multicolumn{1}{c|}{\textbf{55.3 }} & \multicolumn{1}{c|}{\textbf{61.6 }} & \multicolumn{1}{c|}{\textbf{65.8 }} & \multicolumn{1}{c|}{\textbf{69.5 }} & \multicolumn{1}{c|}{\textbf{43.3 }} & \multicolumn{1}{c|}{\textbf{53.2 }} & \multicolumn{1}{c|}{\textbf{59.9 }} & \multicolumn{1}{c|}{\textbf{64.5 }} & \multicolumn{1}{c}{\textbf{69.0 }} \\
\modifyF{LOMO+D-CCA} & 43.3  & 54.3  & 61.2  & 65.3  & 68.9  & 42.4  & 51.8  & 58.1  & 62.9  & 66.9  \\
\modifyF{LOMO+D-SPA} & 44.1  & 54.5  & 60.7  & 65.2  & 68.9  & 41.9  & 51.3  & 57.8  & 62.9  & 67.2  \\
\hline
\end{tabular}%
\label{tableTED}%
\end{table}%

\begin{figure}
\center
\subfigure[3DPeS]{
   \includegraphics[width=0.6\linewidth]{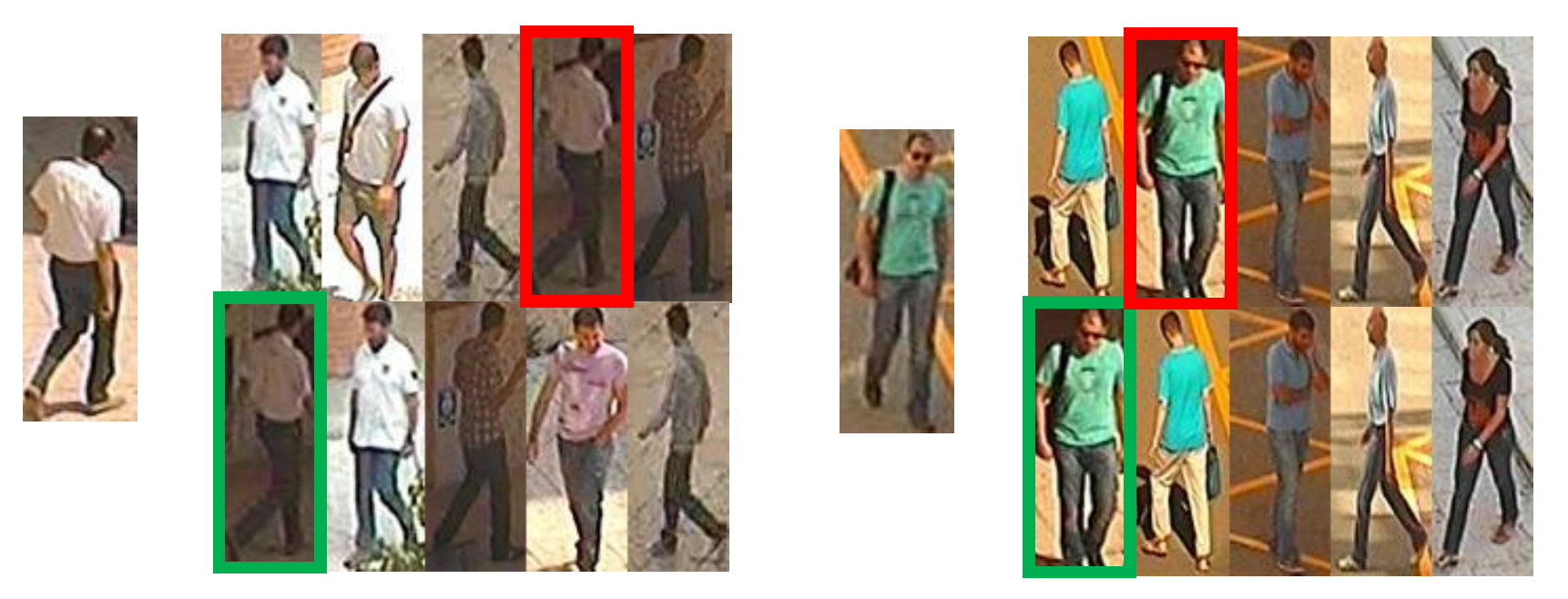}
}
\subfigure[CAVIAR4REID]{
   \includegraphics[width=0.6\linewidth]{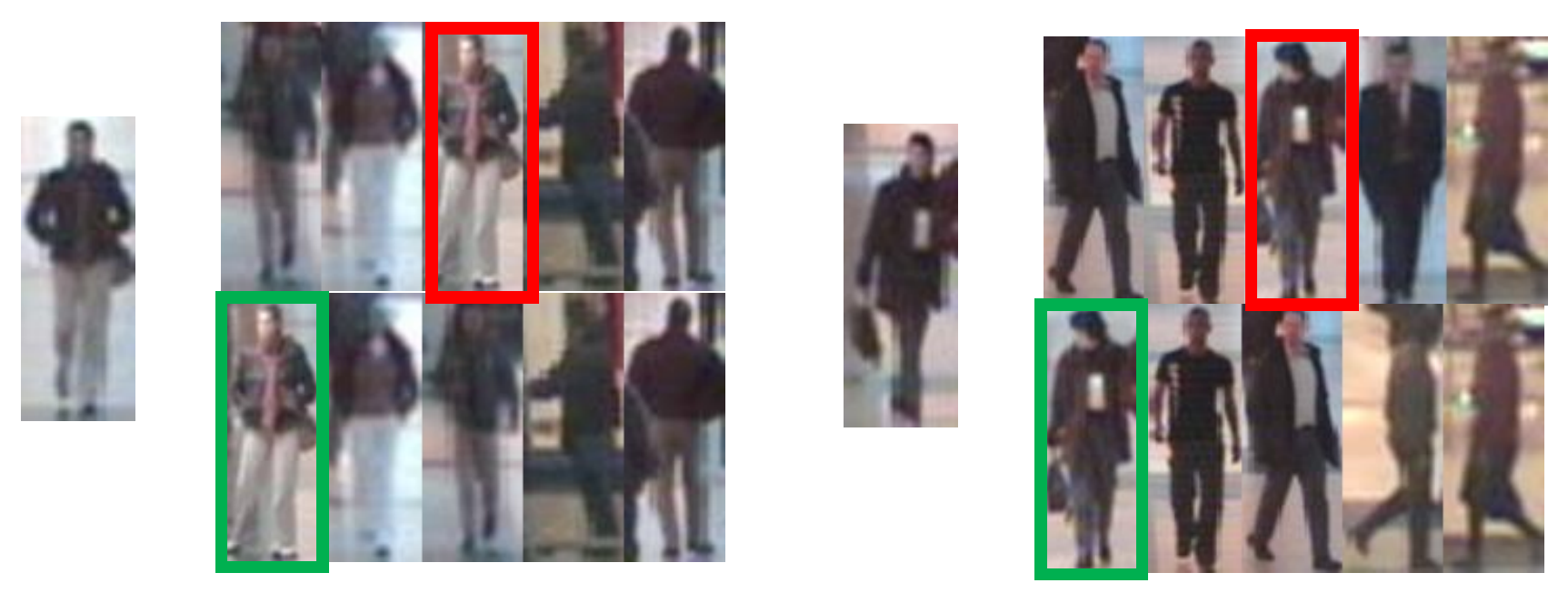}
}
\caption{Top 5 matching gallery images on 3DPeS and CAVIAR4REID. In each group of images, the probe image is on the left. The first and second rows are the matching results of LOMO and LOMO+TED when using MLAPG metric. The bounding boxes show the correct matchings.}
\label{fig:matching}
\end{figure}

To analyze the results more specifically, we also compare some matching samples of LOMO and LOMO+TED when using MLAPG metric in Figure \ref{fig:matching} to see in what situations the transferred depth features would  help to get more robust matching. Among the four groups of images, the two groups on the left show that depth feature can help to find the person with similar body shape, when appearance color changes due to illumination. The other two groups on the right show that, when the appearance color and texture of two pedestrians are very similar, depth feature can help to distinguish the correct matching by body shape. 

\begin{figure*}
  \centering
  \subfigure[ELF18(LFDA)+TED]{
   \includegraphics[width=0.2\linewidth,height=0.2\linewidth]{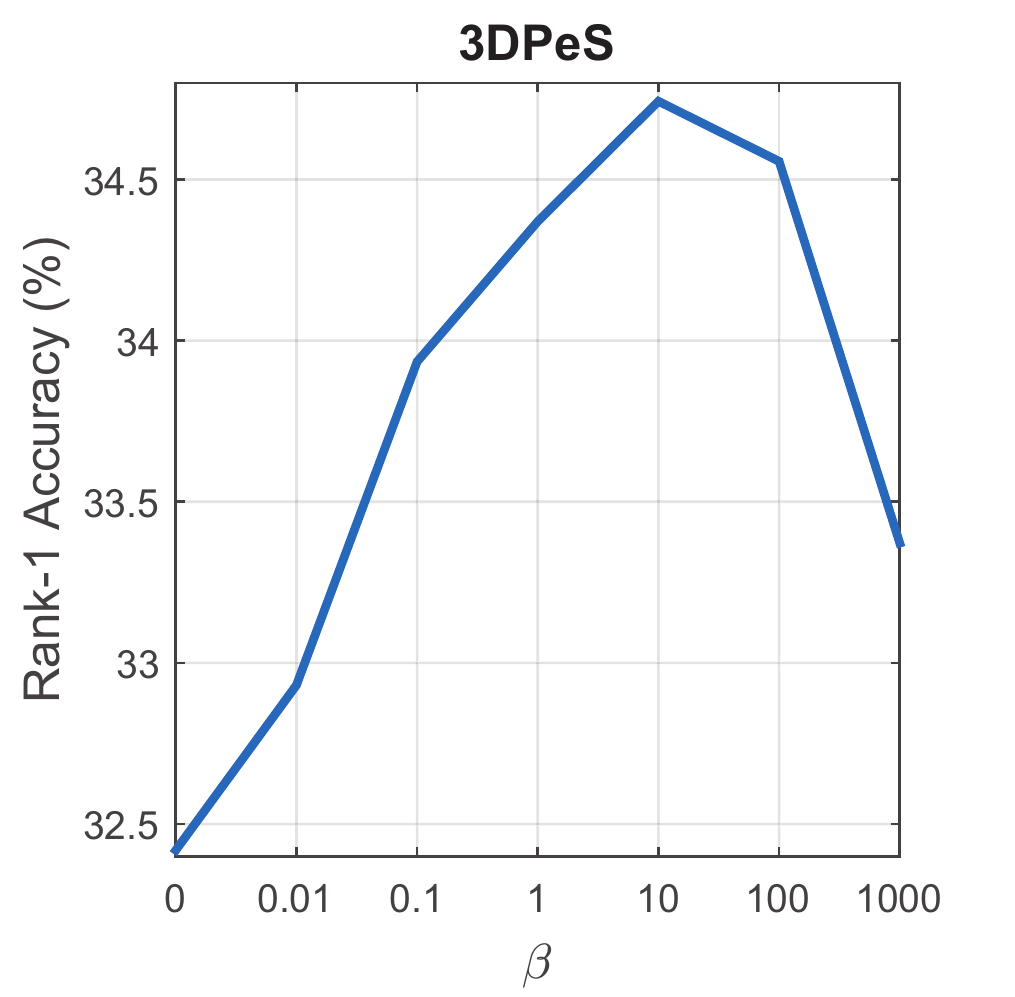}
  }
  \subfigure[LOMO(MLAPG)+TED]{
   \includegraphics[width=0.2\linewidth,height=0.2\linewidth]{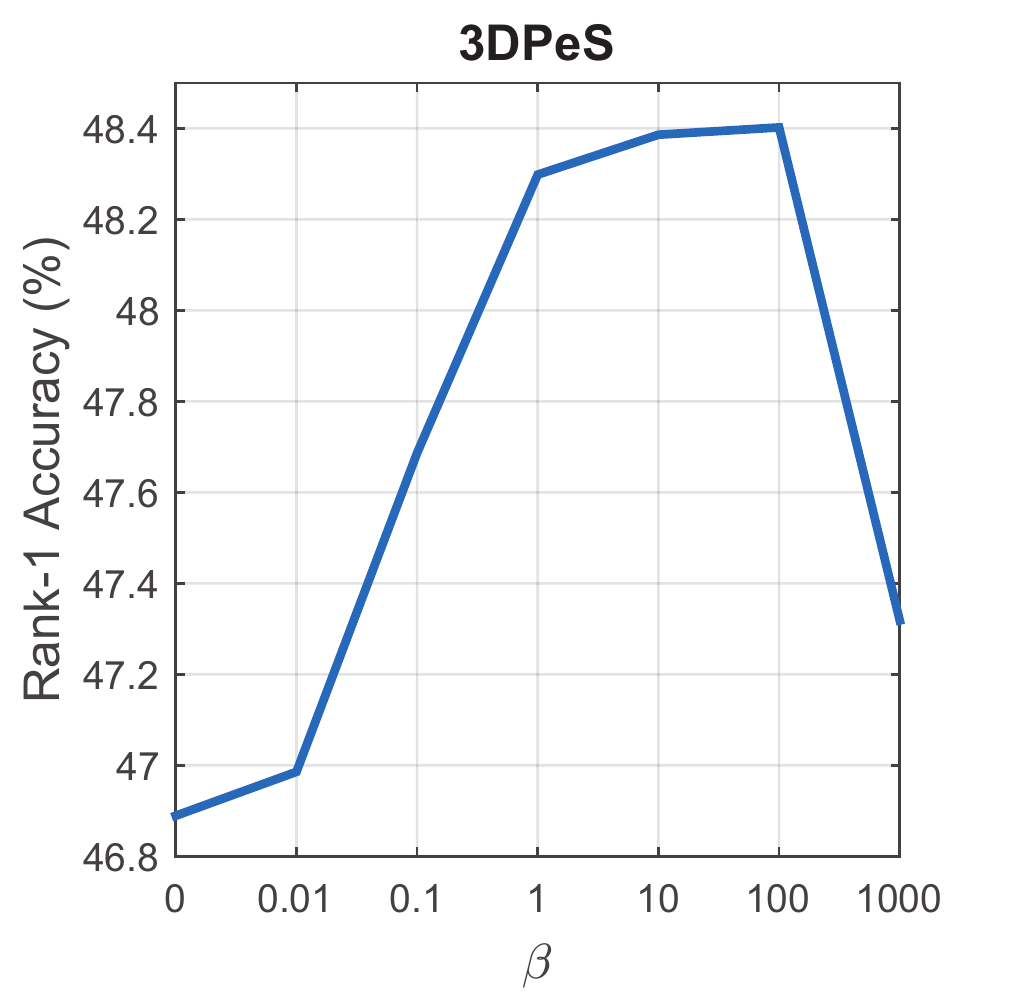}
  }
  \subfigure[ELF18(LFDA)+TED]{
   \includegraphics[width=0.2\linewidth,height=0.2\linewidth]{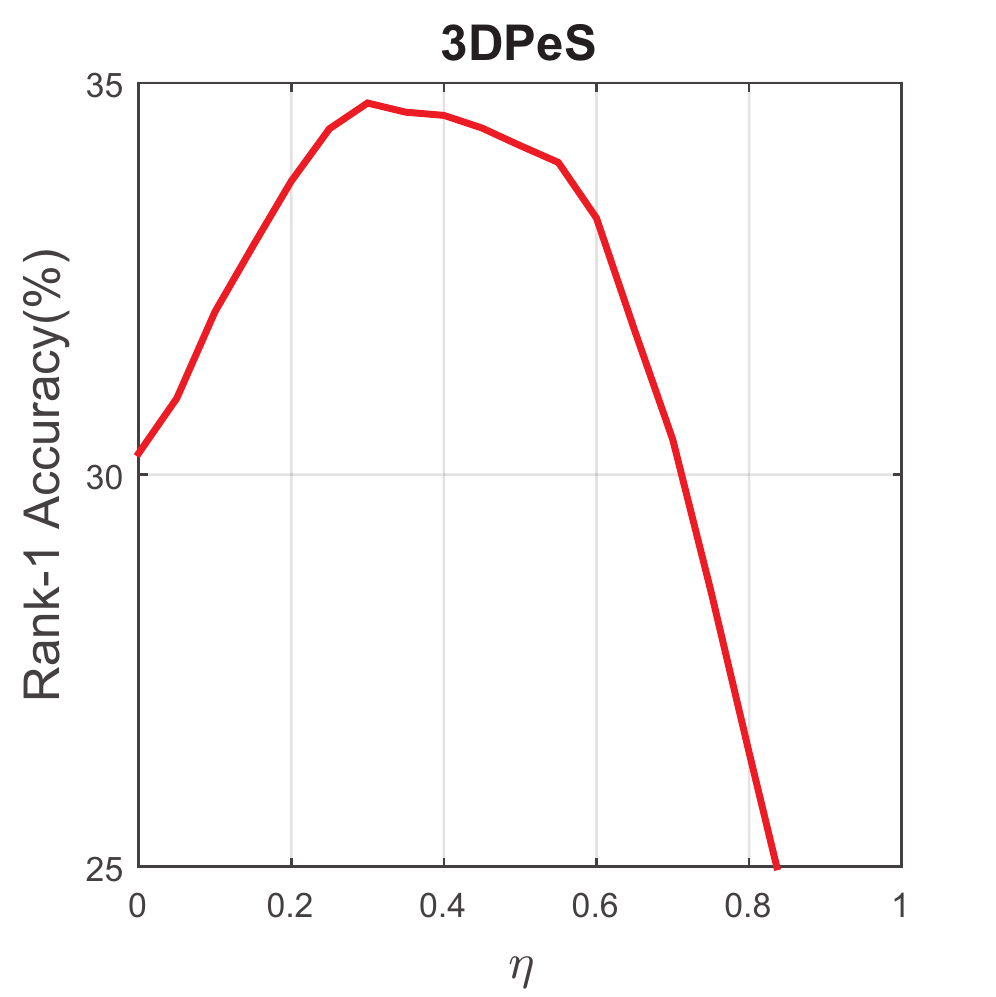}
  }
  \subfigure[LOMO(MLAPG)+TED]{
   \includegraphics[width=0.2\linewidth,height=0.2\linewidth]{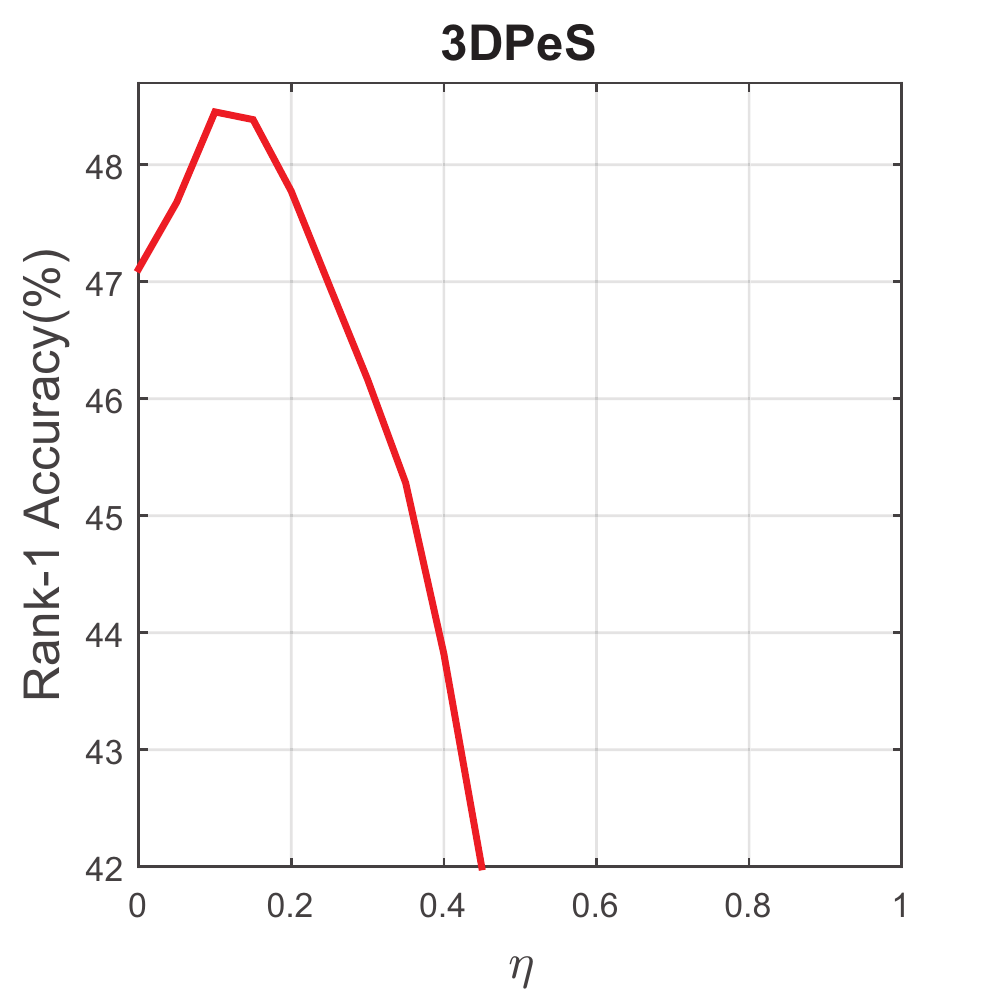}
  }
  \subfigure[ELF18(LFDA)+TED]{
   \includegraphics[width=0.2\linewidth,height=0.2\linewidth]{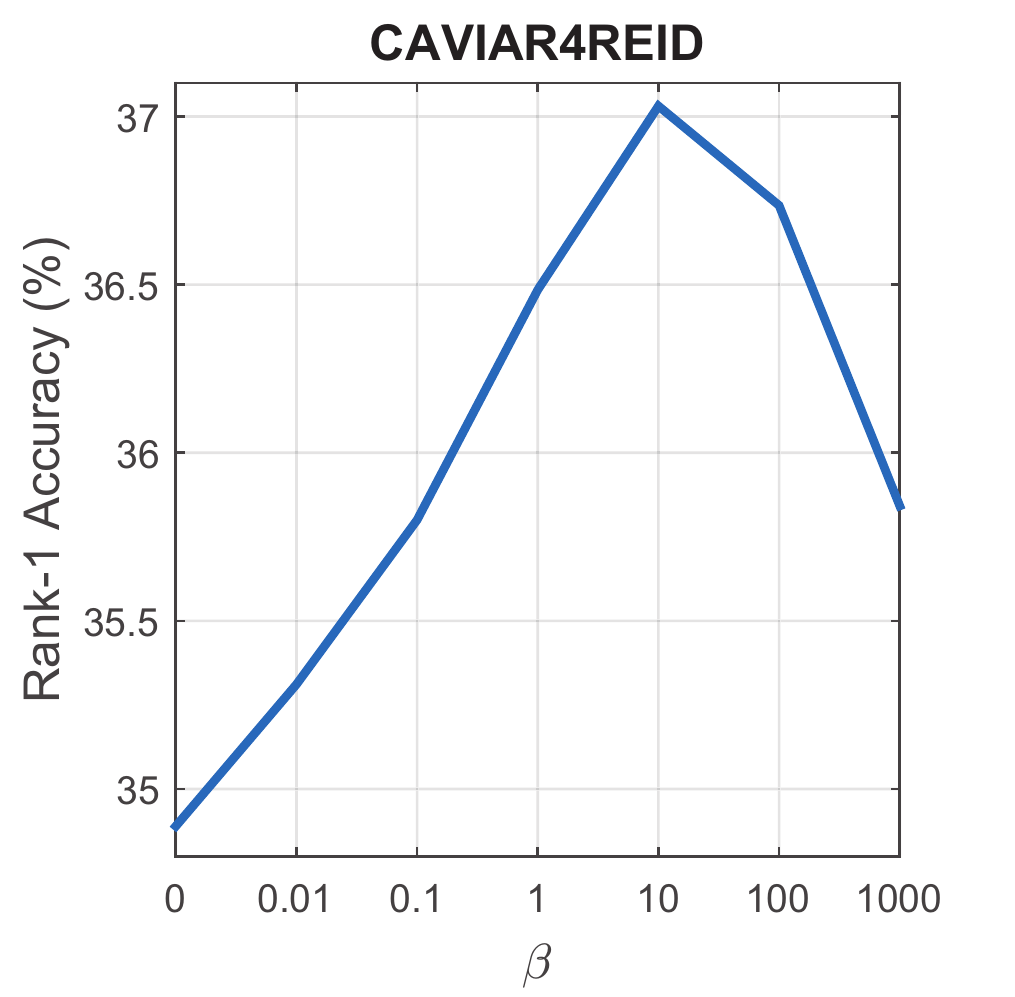}
  }
  \subfigure[LOMO(MLAPG)+TED]{
   \includegraphics[width=0.2\linewidth,height=0.2\linewidth]{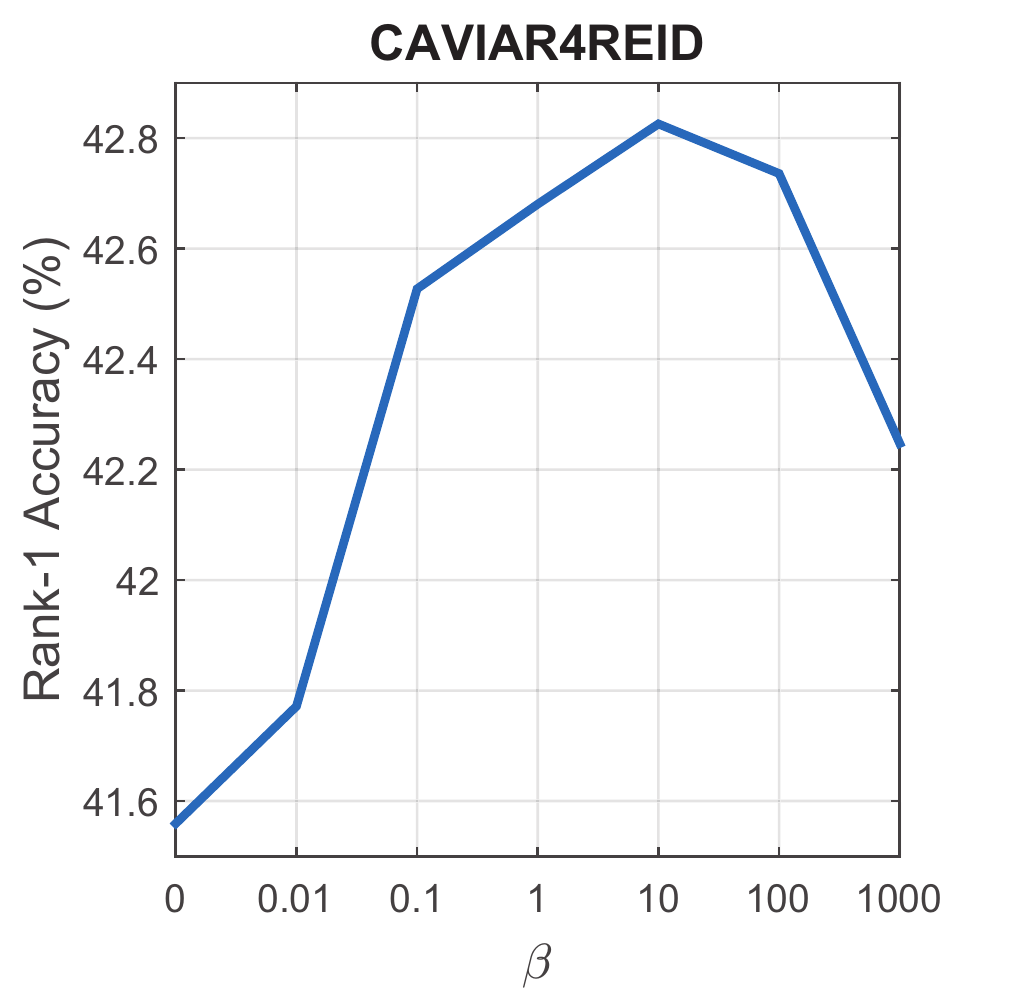}
  }
  \subfigure[ELF18(LFDA)+TED]{
   \includegraphics[width=0.2\linewidth,height=0.2\linewidth]{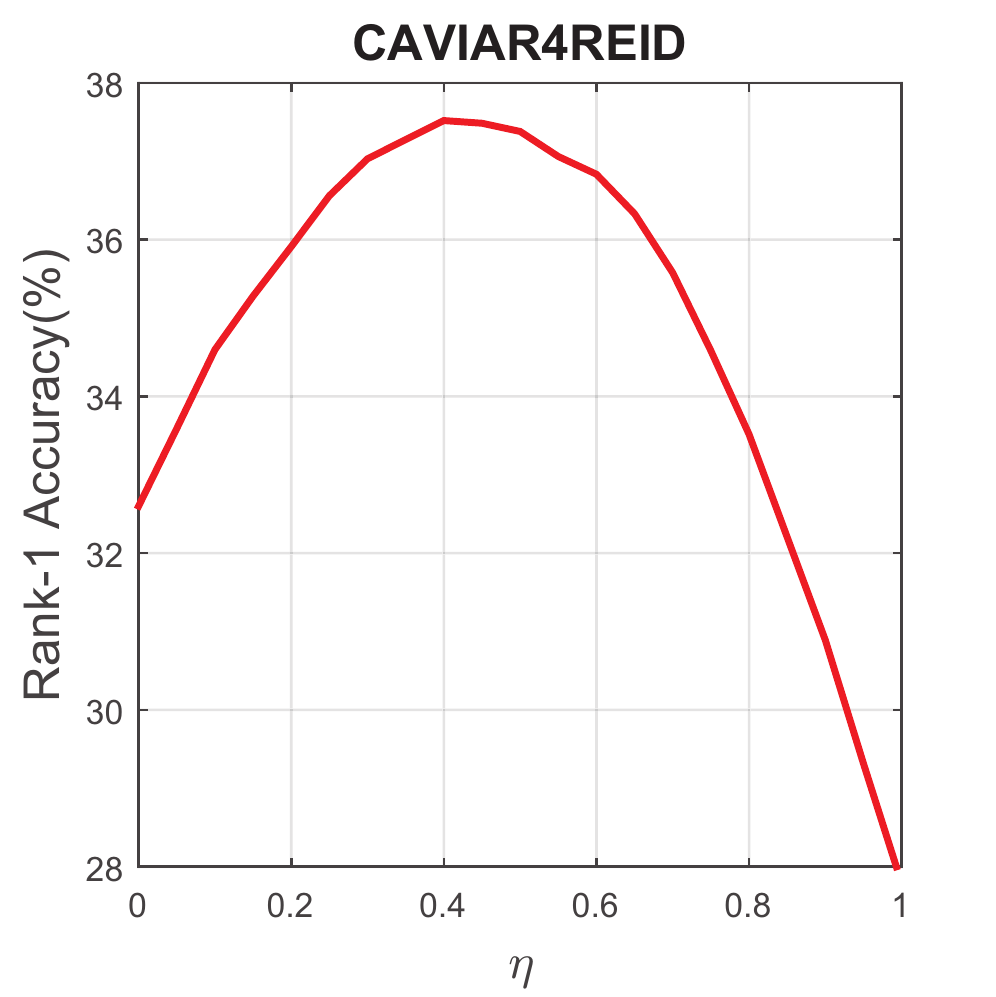}
  }
  \subfigure[LOMO(MLAPG)+TED]{
   \includegraphics[width=0.2\linewidth,height=0.2\linewidth]{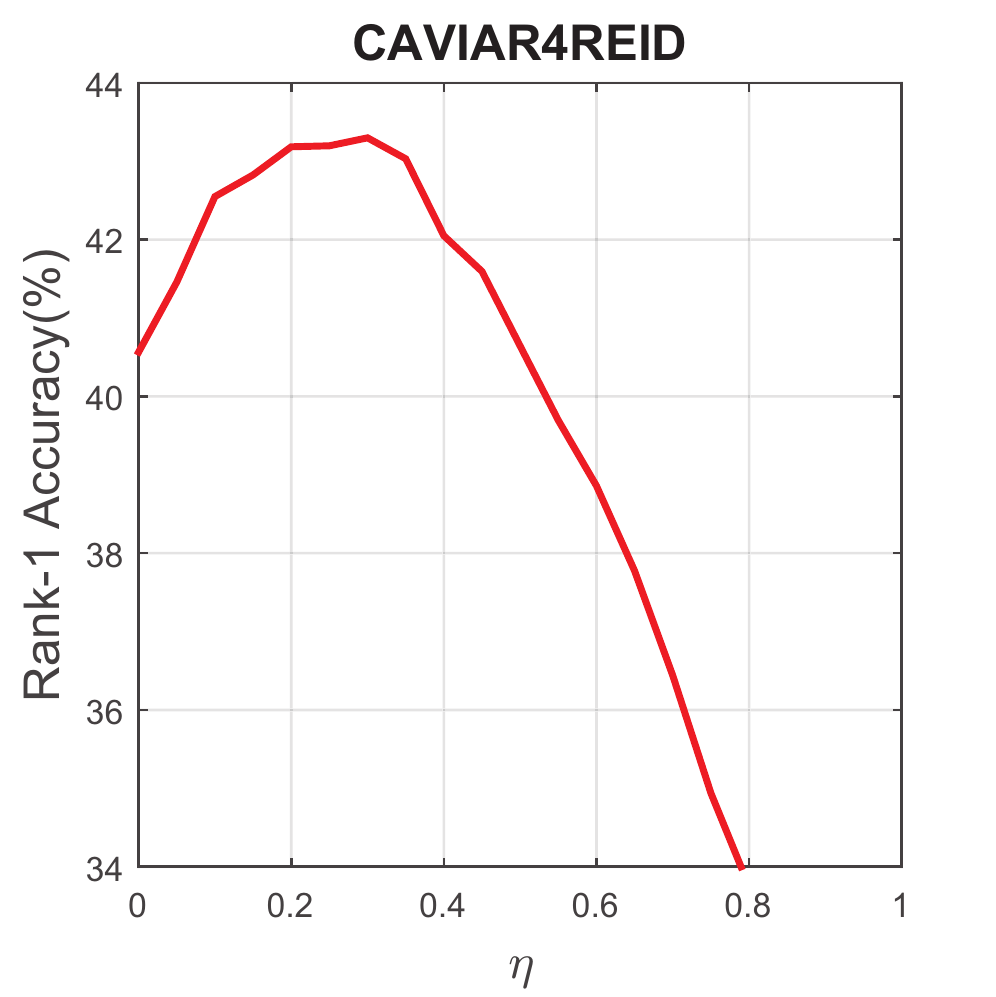}
  }
  \caption{Effects of parameters $\beta$ and $\eta$ (shown by rank-1 accuracy). $\beta$ is the weight of heterogeneous feature mapping term $\Omega_{vd}$ and $\eta$ is the fusion weight of appearance feature and transferred Eigen-depth feature (TED). Two different combinations of appearance features and metrics ELF18(LFDA)+TED and LOMO(MLAPG)+TED on both 3DPeS and CAVIAR4REID are reported here. }\label{fig:para}
\end{figure*}

\vspace{0.1cm}
\noindent \textbf{Effects of Parameters}. To further analyze the effects of key components in our kernelized implicit feature transfer scheme, we also evaluated two significant parameters in our model, the weight $\beta^\prime$ of the heterogeneous feature mapping term $\Omega_{vd}$ in Equation (\ref{equ:obj_fun}) and distance fusion weight $\eta$ in Equation (\ref{equ:dist_fusion}). Let $\beta^\prime=\frac{\beta}{{\rm tr}(\mathbf{B}_{vd})}$. We varied $\beta$ from 0 to 1000 and $\eta$ from 0 to 1 in order to see how the performance changed. When varying these two parameters, other parameters were fixed and set to default values. The rank-1 accuracies under different parameter settings were reported in Figure \ref{fig:para}. Due to space limitation, we report the results of ELF18(LFDA)+TED and LOMO(MLAPG)+TED on 3DPeS and CAVIAR4REID, while the effects of parameters are similar under other settings.

We first analyse the effect of $\beta$. When $\beta$ is around the default value 10, the best performance is achieved. When the feature mapping term is absent (i.e. $\beta=0$)
, the relation between visual features and depth features is not explored. In this case, we can observe that the improvement of TED is minor. The results indicate that the heterogeneous feature mapping term $\Omega_{vd}$ is effective for transferring more effective features by leveraging auxiliary depth information.

Then we analyze the effect of $\eta$. It can be observed that, for ELF18(LFDA)+TED, the performance is improved significantly within range from $\eta=0.2$ to $\eta=0.5$; for LOMO(MLAPG)+TED, the best parameter value is around 0.15. \modifyD{The RGB-based appearance features are more powerful in most common cases when people do not change their clothes and there is no severe illumination change, and TED can help removing ambiguities of top-ranked matchings.}

\subsection{Runtime Performance Evaluation}
\modifyF{
We tested DVCov and Eigen-depth feature on BIWI RGBD-ID and kernelized implicit feature transfer scheme on CAVIAR4REID to compute the computational cost. In Table \ref{tab:runtime}, the time cost for extracting DVCov, Eigen-depth feature and transferring depth feature is reported.
}
\begin{table}
\centering
\modifyF{
  \caption{runtime performance of extracting DVCov, Eigen-depth feature and transferring depth feature.}
    \begin{tabular}{c|c|c}
    \hline
    Method      & Process     & Time (s) \\
    \hline
    \multirow{3}[6]{*}{DVCov} & \tabincell{c}{Computing normals\\(one frame)} & 0.272 \\
\cline{2-3}                & \tabincell{c}{Extracting DVCov\\(one frame)} & 0.038 \\
\cline{2-3}                & \tabincell{c}{Matching DVCov\\(one pair of frames)} & 0.007 \\
    \hline
    \multirow{3}[6]{*}{Eigen-depth} & \tabincell{c}{Extracting Eigen-depth\\(one frame)} & 0.010  \\
\cline{2-3}                & \tabincell{c}{Training LDA\\(on training set)} & 0.946  \\
\cline{2-3}                & \tabincell{c}{Matching Eigen-depth\\(one pair of frames)} & $3.734\times10^{-7}$ \\
    \hline
    \multirow{3}[6]{*}{\tabincell{c}{Depth feature\\ transfer}} & \tabincell{c}{Computing kernel\\(one frame)} & 0.006  \\
\cline{2-3}                & \tabincell{c}{Training\\(on training set)} & 6.869  \\
\cline{2-3}                & \tabincell{c}{Matching\\(one pair of frames)} & $1.478\times10^{-6}$ \\
    \hline
    \end{tabular}%
  \label{tab:runtime}%
  }
\end{table}%

%

\section{Conclusion \& Discussion}

We have developed a depth-based person re-identification model and addressed the bottleneck problem in person re-identification in the cases of clothing change and extreme illumination, which make most existing person re-identification models not workable. In our work, we have proposed two depth shape descriptors: \modifyD{the depth voxel covariance descriptor (within-voxel covariance and between-voxel covariance)} and locally rotation invariant Eigen-depth feature. The local rotation invariance property of Eigen-depth feature has been proven in theory. By combining skeleton-based feature which provides complementary information to our proposed depth shape descriptors, a complete depth-based modeling of a person is formed. Extensive experiments on three RGB-D re-identification datasets show that RGB appearance-based methods suffer from clothing change and extreme illumination, while Eigen-depth feature does not \modifyD{and is able to describe shape} and deal with rotation better than existing methods.

\modifyA{
Since depth information is not always available, we have further proposed a kernelized implicit feature transfer scheme to estimate the Eigen-depth features from RGB images. By augmenting RGB-based appearance feature with the implicitly estimated depth feature, it is helpful for further reducing visual ambiguities for top-ranked matchings and boosting the top rank accuracy in person re-identification.
}

\modifyG{
Finally, we summarize the advantages of our proposed method as follows.
Firstly, our framework quantifies depth information, and this can achieve clearly better performance than applying RGB appearance information in the cases of clothing change and extreme illumination change; secondly, compared to existing 3D shape descriptors, the proposed depth-based features can better describe human body shape in depth images due to the extraction of (logarithmic) Eigen-depth feature, which is locally rotation invariant; thirdly, the proposed depth shape descriptor and the skeleton-based feature are complementary to each other, and the combination of them can provide more discriminant information of human body and thus achieve better performance.

For future development, the self-occlusion and mutual-occlusion problems remain largely unsolved in RGB-based person re-identification, and they are indeed also challenging for our proposed depth-based person re-identification model.
Especially, when a large part of body is occluded, it is much harder to predict the 3D shape of a body. In such a case, the occlusion becomes a typically difficult problem, and more investigation is needed in the future. A possible way may be to consider the partial person re-identification problem as discussed in \cite{Zheng2016partial} for the depth-based approach.
}



\section*{Acknowledgment}

The authors would like to thank the reviewers' constructive comments.

\ifCLASSOPTIONcaptionsoff
  \newpage
\fi



\bibliographystyle{IEEEtran}
\bibliography{depth_reid}

\begin{IEEEbiography}
[{\includegraphics[width=1in,height=1.25in,clip,keepaspectratio]{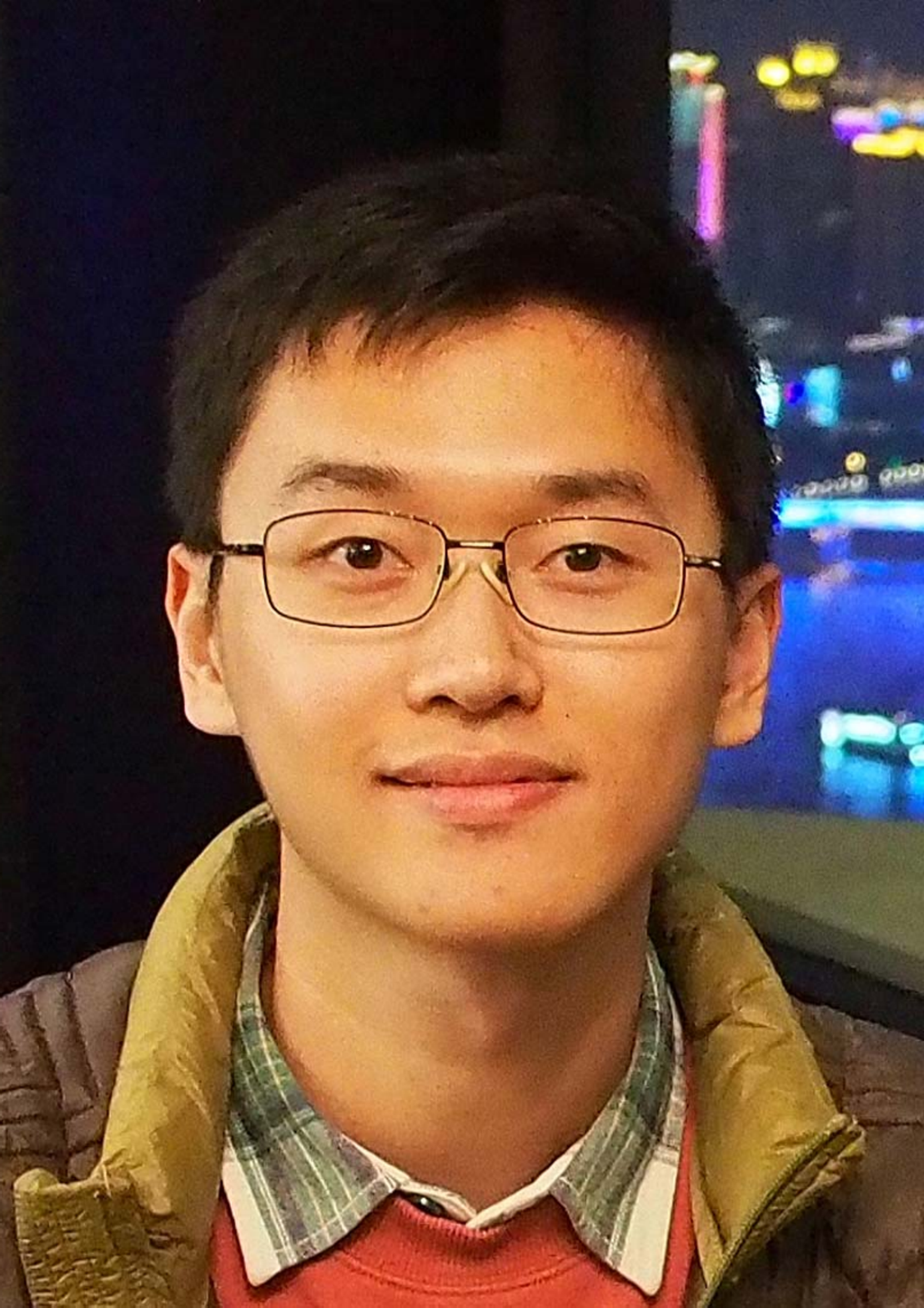}}]
{Ancong Wu}
received the bachelor's degree in intelligence science and technology from Sun Yat-Sen University in 2015.
He is pursuing PhD degree with the School of Electronics and Information Technology in Sun Yat-sen University.
His research interests are computer vision and machine learning. He is currently focusing on the topic of person re-identification. \\ Homepage: \url{http://isee.sysu.edu.cn/~wuancong/}.
\end{IEEEbiography}

\begin{IEEEbiography}
[{\includegraphics[width=1in,height=1.25in,clip,keepaspectratio]{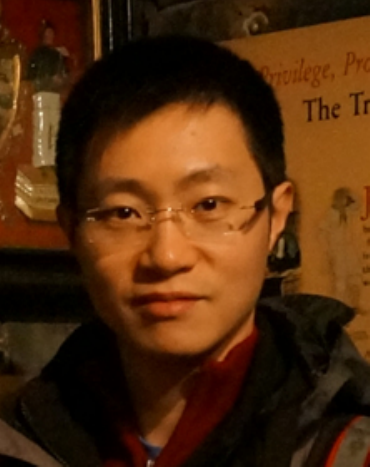}}]
{Wei-Shi Zheng}
received the PhD degree in
applied mathematics from Sun Yat-sen University
in 2008. He is a Professor with Sun Yat-Sen
University. He has been a postdoctoral researcher
on the EU FP7 SAMURAI Project with Queen Mary
University of London. His recent research interests
include person re-identification, action/activity recognition, and large-scale machine learning algorithms. He has joined Microsoft Research Asia
Young Faculty Visiting Programme. He has outstanding reviewer award in ECCV 2016. He is a
recipient of the Excellent Young Scientists Fund of the National Natural Science Foundation of China, and a recipient of Royal Society-Newton
Advanced Fellowship. \\ Homepage: \url{http://isee.sysu.edu.cn/~zhwshi/}.
\end{IEEEbiography}

\begin{IEEEbiography}
[{\includegraphics[width=1in,height=1.25in,clip,keepaspectratio]{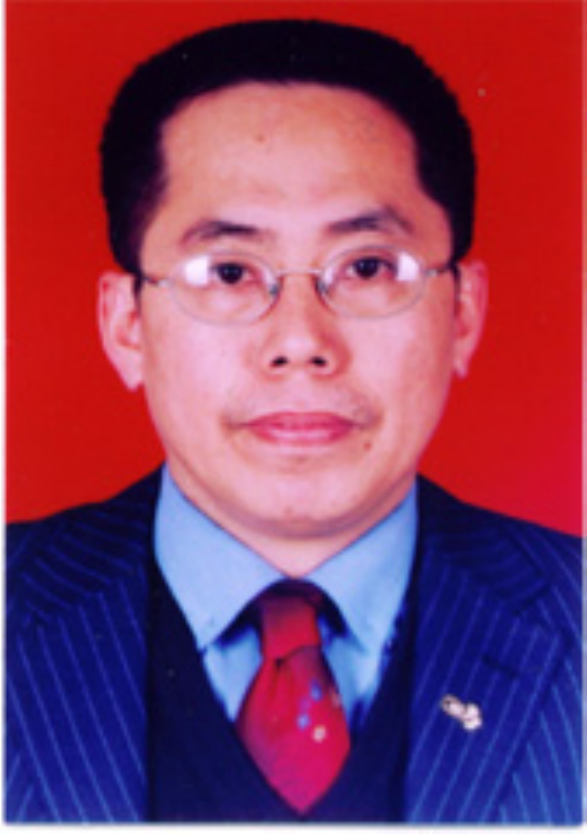}}]{Jian-Huang Lai} received the PhD degree in mathematics from Sun Yat-sen University in 1999. He is a Professor of School of Data and Computer Science in Sun Yat-sen university. His current research interests are in the areas of digital image processing, pattern recognition, multimedia communication, wavelet and its applications. He has published over 100 scientific papers in international journals and conferences including IEEE TPAMI, IEEE TNN, IEEE TIP, IEEE TSMC-B, PR, ICCV, CVPR, and ICDM.
\end{IEEEbiography}

This paper has supplementary downloadable material available at http://ieeexplore.ieee.org, provided by the author. The material includes a document of more experiments and analysis. Contact wuancong@mail2.sysu.edu.cn for further questions about this work.

\end{document}